\documentclass[12pt]{amsart}
\usepackage{amsbsy,amssymb,amsmath,amsthm,amscd,amsfonts,latexsym,amstext,delarray,
amsmath,graphicx} 
\usepackage{qtree}
\usepackage[margin=.8in]{geometry}
\usepackage{color}
\usepackage[all]{xy}
\usepackage{tipa}

\newtheorem{thm}{Theorem}[section]
\newtheorem{prop}[thm]{Proposition}
\newtheorem{cor}[thm]{Corollary}
\newtheorem{lem}[thm]{Lemma}

\newtheorem{defn}[thm]{Definition}
\newtheorem{rem}[thm]{Remark}
\newtheorem{ex}[thm]{Example}

\usepackage[normalem]{ulem}

\usepackage[]{mdframed}

\usepackage{bigstrut}

\usepackage[utf8]{inputenc}
\usepackage{graphicx,hyperref}
\usepackage{subfiles}
\usepackage{wrapfig}
\usepackage{amsmath}
\usepackage{amssymb}
\usepackage{paracol}
\usepackage{multicol}
\usepackage{adjustbox}
\usepackage[inkscapeformat=png]{svg}

\usepackage[square,numbers]{natbib}
\usepackage[utf8]{inputenc}
\usepackage{xcolor}
\usepackage{soul}
\usepackage{tikz}
\usepackage{latexsym,amscd,amssymb,epsfig}
\usepackage{caption}
\usepackage{pgfplots}
\usepackage{pgfplotstable}
\pgfplotsset{compat=1.7}
\calclayout
\usetikzlibrary{calc}
\usetikzlibrary {positioning}
\usepackage[linguistics]{forest}
\usepackage{bbm}
\usepackage{mathtools}
\forestset{
  nice empty nodes/.style={
    for tree={calign=fixed edge angles, calign angle=50},
    delay={where content={}{shape=coordinate,
    for current and siblings={anchor=north}}{}}
  }
}

\numberwithin{equation}{section}

\def\Q{{\mathbb Q}}

\def\R{{\mathbb R}}

\def\cC{{\mathcal C}}

\def\cF{{\mathcal F}}

\def\cK{{\mathcal K}}
\def\cL{{\mathcal L}}
\def\cM{{\mathcal M}}
\def\cN{{\mathcal N}}
\def\cO{{\mathcal O}}
\def\cP{{\mathcal P}}

\def\cS{{\mathcal S}}

\def\cV{{\mathcal V}}
\def\cW{{\mathcal W}}

\def\bI{{\mathbb I}}

\def\bO{{\mathbb O}}

\def\Hom{{\rm Hom}}

\def\fM{{\mathfrak M}}
\def\fT{{\mathfrak T}}
\def\fF{{\mathfrak F}}
\def\fO{{\mathfrak O}}
\def\fB{{\mathfrak B}}

\def\fC{{\mathfrak C}}
\def\fP{{\mathfrak P}}

\title[Morphology-Syntax Interface]{The Algebraic Structure of Morphosyntax}
\author{Isabella Senturia and Matilde Marcolli}
\date{\today}

\address{Department of Linguistics, 
Yale University, New Haven, CT 06511, USA}
\email{isabella.senturia@yale.edu}

\address{Department of Mathematics and Department of Computing and Mathematical Sciences, 
California Institute of Technology, Pasadena, CA 91125, USA}
\email{matilde@caltech.edu}

\begin{document}

\begin{abstract}
    Within the context of the mathematical formulation of Merge and the Strong Minimalist Thesis, 
    we present a mathematical model of the morphology-syntax interface. In this setting, morphology
    has compositional properties responsible for word formation, organized into a magma of morphological
    trees. However, unlike syntax, we do not have movement within morphology. A coproduct decomposition
    exists, but it requires extending the set of morphological trees  beyond those which are generated solely by the magma, to a larger set of possible morphological
    inputs to syntactic trees. These participate in the formation of 
    morphosyntactic trees as an \textit{algebra over an operad}, and a \textit{correspondence between algebras over an operad}.
    The process of structure formation for morphosyntactic trees can then be described in terms of this operadic 
    correspondence that pairs syntactic and morphological data and the morphology coproduct. 
    We reinterpret in this setting certain operations of Distributed Morphology as transformation that
    allow for flexibility in moving the boundary between syntax and morphology within the morphosyntactic objects.
\end{abstract}

\maketitle

\section{Introduction} \label{IntroSec}

This work furthers the development of a mathematical formulation for generative linguistics,  as initiated in \cite{marcolli_mathematical_2024} in the context of Chomsky's Strong Minimalist Thesis, by extending the mathematical formulation of Merge and the
core computational structure of syntax to another component of the language faculty---the morphology-syntax interface. 

The morphology-syntax interface is linguistically conceptualized as how the internal structure of words relates to structures generated by syntax, and the extent to which the rules generating the former correlate to the rules constraining the latter \cite{embick_distributed_2007}. We also interpret it as the ways that syntactic and morphological structures and processes combine in the assembly of morphosyntactic objects.

In order to explore the interface of syntax and morphology mathematically, the mathematical structure of morphology must first be established. The two prominent perspectives on morphology are Nanosyntax and Distributed Morphology (DM). 

Nanosyntax \citep{caha_nanosyntax_2009,starke_nanosyntax_2009} stems from cartography and takes the perspective that the operations underlying syntax also underlie morphology. There is no distinct morphological system: morphological structures are the product of syntactic Merge. With respect to assembly of morphological components, there is a spellout loop comprised of syntactic assembly of a tree up to completion of a phase, then lexicalization of the section of the syntactic structure, before more syntactic structure-building and more spellout in the next phase, and so on. 
On the other hand, Distributed Morphology \cite{halle_distributed_1993,halle_key_1994} views morphology as the housing of morphological features in leaves of syntactic trees which can be manipulated by a variety of operations. 

In this work, we take an intermediary perspective---morphological features end up as feature bundles in leaves of syntactic trees, but these bundles are actually hierarchical tree structures with morphological features at the leaves and feature bundles at internal nodes. The model we present shares with Nanosyntax the idea that the underlying fundamental algebraic operations in morphology rely on the same computational structure that governs syntax, but with some specific adaptations that control and manipulate the feature hierarchies and the flexible 
boundary between morphology and syntax.

To avoid possible misunderstandings of our approach as presented in this paper,
we begin by providing some general guidelines for how to interpret the construction
presented in this paper. A presentation more focused on the linguistic interpretation
and less detailed on the mathematical properties will be presented as a separate
forthcoming paper, which will also help clarify some of these interpretations.
Some important aspects to keep in mind while reading through the paper
are listed here.
\begin{itemize}
\item {\em Our use of syntax, morphology, and morphosyntax terminology.} By syntax we mean the
fundamental computational structure based on free symmetric Merge acting on workspaces, in the
mathematical formulation developed  \cite{marcolli_mathematical_2024}. By morphology we simply mean
here the assembly of feature bundles, sometimes known in morphology as the theory of bundling. The underlying algebraic structure is fully inherited from syntax
(it is the same free non-associative commutative magma operation underlying the formation of syntactic
objects), and the only difference is the labeling procedure: while in syntax this is determined by a head
function, and is a separate mathematical function distinct from the magma, a simpler union operation suffices for feature bundle formation, and the magma itself applies the label. 
By morphosyntax we then mean combination (via a geometric correspondence) of the 
computational structure of syntax and the feature bundles. 

\item {\em Computation is governed by syntax.} The mathematical model of Merge developed in 
\cite{marcolli_mathematical_2024} shows that a very constrained algebraic structure describes
the structure formation via free symmetric Merge. The two key components of this structure are
the magma operation and a coproduct operation that allows for the extraction of accessible terms
as computational material. Other parts of the faculty of language, such as the syntax-semantics
interface, also discussed in \cite{marcolli_mathematical_2024}, are syntax-driven, and rely on
these same computational mechanisms. The situation with morphology is similar: again 
the magma operation and the coproduct operation that exist in syntax are the key
computational structures (in this sense, it is \textit{syntax all the way down}), and the interface
relies then simply on a correspondence matching lexical items and syntactic features  
to feature bundles. However, this correspondence does not mean that the syntactic tree assembly is not occurring with morphological objects at the leaves---rather, one can conceptualize the lexical items and syntactic features as being ``placeholders'' for the morphological objects. The existence of the correspondence does \textit{not} mean that with respect to the timing of the derivation steps, the derivation is happening with objects that are not morphological in nature. The next point discusses this notion of timing of derivation versus computational components further.

\item {\em Decoupling of independent computational substructures versus derivational steps.}
A key idea in Chomsky's recent (starting around 2013) formulation of Minimalism is the
decoupling of all parts of the computational structure that can be decoupled. This idea led
directly to identifying a free symmetric Merge as the core computational mechanism, interfaced
with a filtering system (for well formed structure, in terms of phases, theta role assignments),
necessary for parsing at the syntax-semantics interface, 
and further language specific filters at Externalization (parametric variation). 
In this description of syntax, one can take two different, but provably equivalent, viewpoints. The key to understanding this fact is that the separation of 
independent computational components is not a separation in terms of time ordering of
operations in a derivation, but rather a separation of their conceptual algebraic structures. 
This decoupling of independent computational modules has the advantage of rendering
the algebraic structure transparent and making it easy to prove general results (which help to
reduce the number of assumptions of the model and select between alternatives). On the
other hand, efficiency of algorithmic realizations prefers a viewpoint in which the filtering
is done step by step along structure formation. While this may appear at odds with the
decoupling of structure formation and filtering, in fact these two perspective are
fully equivalent (this equivalence is proven in \cite{marcolli_theta_2025} and \cite{marcolli_phases_2025}). 
The situation we present here in describing morphology and what we refer to as the
morphology-syntax interface (and morphosyntax) is an analogous situation, where
for the sake of clarity in the identification of the key algebraic structures, we decouple
all the substructures that can be decoupled and describe them individually along
with their interface mechanism. This again should not be read as a separation,
at the level of time-ordered sequences of derivations, between syntax and morphology: there is no such thing, as the core computational structure is always driven by syntax.

\item {\em Operations of Distributed Morphology.} Finally, the algebraic structures
we identify provide a mathematical formulation for the fusion, fission, obliteration and 
impoverishment operations of Distributed Morphology. In view
of the nature of the core computational mechanism (the magma and coproduct
operations), this shows that it is fusion and fission that form the key basic
operations. Fusion and fission switch between the two modes of labeling of the (syntactic) magma, unlabeled (syntax) and labeled (morphological feature bundle formation), and therefore flexibly moving the 
syntax-morphology boundary, that is, the boundary between the unlabeled syntactic nodes and the labeled, by either individual features or feature bundles, morphological nodes. The obliteration and impoverishment operations,
which are often presented as the fundamental ones in Distributed Morphology,
are more naturally seen (and mathematically provable) as derived from the fusion/fission ones and
the coproduct. Not only is this a striking result from the morphological perspective, but mathematically it again demonstrates a reduction of the number of required operations and a simplification of the overall computational structure of the system. 
\end{itemize}

As with the syntactic objects, the morphological tree structures are generated by a magma operation. The two algebraic structures of
syntax and morphology begin to differ at the level of workspaces and their algebraic structure. In syntax, in the model of 
 \cite{marcolli_mathematical_2024}, workspaces are forests of syntactic objects, and the vector space they span is endowed
 with a Hopf algebra structure, with the action of Merge realized as a Hopf algebra Markov chain. In the case of morphology,
 one can also form workspaces with a coproduct operation that allows for extraction and elimination of parts of
 feature bundles, but this operation requires an extension of the underlying space beyond the objects formed by the magma, which is obtained via a comodule structure.
 The morphological objects obtained in this way then provides the inputs at the leaves of syntactic trees. This insertion of morphological
 structures at the leaves of syntactic trees is realized algebraically as an algebra over an operad, as suggested in 
 \cite{marcolli_mathematical_2024} and in \cite{marcolli_extension_2024}, but with a subtlety: we need to introduce a more refined notion of {\em correspondence between algebras over an operad}. 
 Moreover, a main difference with respect to syntax is that,
 instead of having a Merge operation acting as a Hopf algebra Markov chain on the morphological workspaces, as with the syntactic space, one
 has a more complex structure formation operation that acts only at the level of morphosyntactic trees. Since in
 these trees the boundary between syntax and morphology is somewhat flexible (in a way that is made precise by
 two operations, fusion and fission, that push the boundary either upward or downward, respectively), this action can sometimes be interpreted in
 syntactic or in morphological terms. 
 
 Thus, the morphology-syntax interface in this work is described by a number of mathematical systems 
 operating partly sequentially and partly in parallel and interfacing with each other, whose mathematical 
  structures are not identical but must be compatible with each other to the extent that they must compose 
  to form, and operate on, the final morphosyntactic structures. It should be pointed out that this entire model
  is pre-Externalization. Externalization then acts on the structures obtained in this way as a selection process
  governed by {\em morphosyntactic parameters} that select viable structures in a language-dependent way
  and incorporates language-specific lexicon. Then, in parallel to the process in syntax, any ill-formed structures 
  will be filtered out via morphosyntactic parameters. One of these will be compatibility with Feature Geometry \cite{harleyPersonNumber2002}. 
  We will not discuss the Externalization part and the parametric
  variation in this paper, where we only focus on the underlying computational structure that acts in a way
  that is independent of realization in a specific language. 
  
  The following is a quick summary of the various components of the Morphology-Syntax interface, 
  and their consecutive order.
  
 \begin{enumerate}
 \item Syntactic and morphological workspaces exist simultaneously and independently:
\begin{enumerate}
    \item {\em Syntax:} The workspaces are forests whose components are syntactic objects (the elements of
    the free non-associative commutative magma generated by the set of lexical items and syntactic features).
    The linear span of workspaces has a coproduct that extracts accessible terms for structure formation and movement
    realized by Merge. The Merge action on workspaces has the form of a Hopf algebra Markov chain with respect to the coproduct
    that extracts accessible terms and the product that forms workspaces.
    This is the mathematical model of Merge developed in \cite{marcolli_mathematical_2024}: it will be briefly 
    summarized in the following section.
    \item {\em Morphology:} Morphological trees are also built by a free non-associative commutative magma, 
    as binary tree structures with morphological features (some of which are valued, some of which are not) at the leaves.
    Workspaces now include components that are not just the elements of this magma (unlike syntax) but include
    additional objects generated by extraction and elimination of features via the coproduct. The resulting linear space of
    these extended morphological workspaces then has a resulting Hopf structure with product and coproduct. 
\end{enumerate}
\item {\em Morphosyntactic trees} are obtained as insertion of morphological trees at the leaves of syntactic
trees in a way that satisfies a compatibility rule between the syntactic features labeling the leaves of the syntactic trees and the morphological
feature bundles labeling the roots of the morphological trees.  Syntactic objects form an algebra over an operad
(as shown in \cite{marcolli_mathematical_2024} and in \cite{marcolli_theta_2025}), and the correct mathematical
structure that describes this insertion of morphological trees into syntactic trees is identified as 
a correspondence of algebras over an operad.
\item {\em Morphosyntactic workspaces} are forests whose components are 
morphosyntactic trees, where the morphological structures inserted at the leaves of the syntactic trees also include
components of morphological workspaces that are not in the magma of morphological trees. The syntactic trees can then be understood as
defining a family of operations that map morphological to morphosyntactic workspaces. These operations are formally
similar to the Hopf algebra Markov chain of Merge acting on syntactic workspaces, but rely on already built syntactic objects. 
\item {\em Distributed Morphology:} post-syntactic morphological operations of fission, fusion, impoverishment, and obliteration 
now can be seen as transformations of morphosyntax trees. They are ``post-syntactic" in the sense that they act after
syntactic trees are interfaced with morphological trees in the formation of morphosyntactic objects, but are still acting prior to
Externalization.
\end{enumerate}

While we recognize that DM is not universally agreed upon as the computational mechanism underlying morphology, the purpose of including this here is to demonstrate that the system is capable of handling post-syntactic morphological operations. Moreover, DM is in particular interesting as it has the overall effect of moving the boundary between syntax and morphology in the morphosyntactic trees.

In addition to these aspects of the model that we will be developing in detail in the rest of the paper,
there are other directions, which we will not be including here but that we expect will also have a 
mathematical formulation compatible with this model. These include:

\begin{itemize}
\item {\em Agree:} valuation of unvalued features in the morphological trees via another colored operad similar to
the treatment of theta theory and phases in \cite{marcolli_theta_2025} and \cite{marcolli_phases_2025}.
\item {\em Externalization: Planarization and Filtering:} in the model we develop here all trees are non-planar. The choice of a planar structure is part of Externalization (and is language-dependent, for example in the structure of prefixes and subfixes in word
formation) as in the case of the planarization of syntactic trees. In the same Externalization phase some of the freely
formed structures are filtered out (again in a language-dependent way) according to morphosyntactic parameters.
\item {\em Externalization: Vocabulary insertion:} after all morphological features have attained their final values, and the morphosyntax trees are in their final configurations, as part of the Externalization process language-specific morphemes are inserted at the leaves of the morphosyntax trees via a colored operad, and word formation takes place according to the features and the structure of the morphological trees. 
\end{itemize}

 In the next sections, we will elaborate on each of the individual components of the pre-Externalization model
 listed above. 
 
 \subsection{Recalling the mathematical structure of syntax}
 
 In \cite{marcolli_mathematical_2024} a mathematical framework for syntax was developed, 
 where the main algebraic structure is a Hopf algebra coproduct, responsible for the extraction
 of accessible terms from syntactic objects that is needed for the core generative procedure 
 underlying the compositional structure of syntactic trees, namely the Merge operation. 
 A brief summary of how Merge operates can be articulated in the following way:
\begin{itemize}
\item Syntactic objects are built by successive iterations of a non-associative commutative binary operation $\fM$,
starting from a finite set $\cS\cO_0$ of lexical items and syntactic features. The resulting set $\cS\cO$ of
lexical items is the free non-associative commutative magma generated by $\cS\cO_0$,
$$ \cS\cO={\rm Magma}_{na,c} (\cS\cO_0,\fM) $$
and as such it can be identified with the set $\fT_{\cS\cO_0}$ of non-planar full binary rooted trees with
leaves decorated by elements of $\cS\cO_0$.
    \item Merge is a dynamical system acting on workspaces: it takes a workspace as input and it outputs a 
    sum of possible resulting workspaces (all the structures obtainable from the input in a single Merge move).
    Structure formation and movement are achieved by iteration of this Merge action. At each step, the current
    workspace that Merge is acting on is likened to a scratchpad, where steps of derivations take place, transforming
    the workspace, starting with an unstructured collection of lexical items, until a final completed sentence
    structure is obtained. At each step a workspace is a disjoint union (a forest) $F=\sqcup_a T_a$
    whose components $T_a\in \cS\cO$ are syntactic objects. 
    \item The vector space $\cV(\fF_{\cS\cO_0})$
    spanned by this set  of forests (workspaces) $\fF_{\cS\cO_0}$ has a product operation $\sqcup$
    that combines two workspaces into a single one (and in particular places syntactic objects into a workspace)
    and a coproduct operation that extracts all the available material for computation in a workspace, namely all
    the accessible terms $T_v \subset T$ of components of the workspace, with $T_v$ the full subtree of $T$ below
    one of the vertices. The coproduct takes the form
    \begin{equation}\label{coprod}
    \Delta(T) = \sum_{\underline{v}} F_{\underline{v}} \otimes T/F_{\underline{v}} \, ,
    \end{equation}
    and with $\Delta(F)=\sqcup_a \Delta(T_a)$ for $F=\sqcup_a T_a$, 
 where $\underline{v}=\{ v_1, \ldots, v_r \}$ are vertices with non-overlapping accessible terms $T_{v_i}$ and
 $F_{\underline{v}}=T_{v_1}\sqcup \cdots \sqcup T_{v_r}$ is the extracted material, with $T/F_{\underline{v}}$
 the corresponding cancellation of the deeper copies. 
    \item the Merge action on workspaces is then formulated as a composition of operations 
    \begin{equation}\label{MergeSS}
\fM_{S,S'} =\sqcup \circ (\fB\otimes {\rm id}) \circ \delta_{S,S'} \circ \Delta, 
\end{equation}    
    where (reading from right to left with $\circ$ indicating composition of functions) 
\begin{enumerate}
    \item first the coproduct $\Delta$ extracts all accessible terms making them 
    available to be used for structure formation.
    \item then $\delta_{S,S'}$ searches over all the extracted terms in the coproduct for a
    specific pair of syntactic objects $S$ and $S^\prime$ to act on, and eliminates all
    terms that are not of the form $S\sqcup S'\otimes F'$ for some forest $F'$.
    \item the remaining terms are acted upon by the operator $(\fB\otimes {\rm id})$
    resulting in terms of the form $\fB(S\sqcup S')\otimes F'$, where the operator $\fB$
    grafts a forest to a common root
    $$ \fB : S \sqcup S' \mapsto \Tree[ $S$ $S'$ ] $$
    \item finally  $\sqcup$ reassembles the new workspace $\fB(S\sqcup S')  \sqcup F'$.
 \end{enumerate}  
 \item this action of the Merge operations $\fM_{S,S'}$ of \eqref{MergeSS} gives rise to three
 possible cases: External Merge, Internal Merge, and Sideward Merge.
 \begin{enumerate}
 \item {\em External Merge:} $\fM_{S,S'}$ where $S$ and $S'$ are syntactic objects that are 
 connected components of the workspace, $F=S \sqcup S' \sqcup \hat F$, resulting in a 
 new workspace $\fM_{S,S'}(F)=\fM(S,S') \sqcup \hat F$ (where $\fM(S,S')=\fB(S \sqcup S')$).
  \item {\em Internal Merge:} $\fM_{S, T/S}\circ \fM_{S,1}$ where $1$ is the unit of the magma (the formal empty tree)
  where $S$ is an accessible term $S=T_v$ of a component $T$ of the workspace $F=T\sqcup \hat F$. 
  Here $\fM_{S,1}$ has the effect
  of extracting $T_v$ and placing it in the workspace and $\fM_{S, T/S}$ then merges it with the remaining structure $T/S$
  resulting in a new workspace $\fM(T_v,T/T_v) \sqcup \hat F$.
 \item {\em Sideward Merge} has three cases: (a) $S=T_v\subset T$ an accessible term and $S'=T'$ a component of
 the workspace $F=T\sqcup T' \sqcup \hat F$, resulting in $\fM(T_v,T') \sqcup T/T_v \sqcup \hat F$; (b) $S=T_v\subset T$
 and $S'=T_w\subset T$ two disjoint accessible terms of the same component $T$, resulting in $\fM(T_v, T_w) \sqcup T/(T_v\sqcup T_w) \sqcup \hat F$; (c) $S=T_v\subset T$ and $S'=T'_w\subset T'$ accessible terms of two different components, resulting in
 $\fM(T_v, T'_w) \sqcup T/T_v \sqcup T'/T'_w \sqcup  \hat F$.
 \end{enumerate}
 \item External Merge and Internal Merge satisfy various cost optimization measures, while Sideward Merge
 is non-optimal, with a hierarchy of non-optimality among the different cases, see \cite{marcolli_extension_2024}.
 \item when all the possible choices of syntactic objects $S$ and $S^\prime$ are considered,
 the Merge operation can be assembled into a single $\cK=\sum_{S,S'} \fM_{S,S'}$ (which despite
 looking like an infinite sum always results in a finite sum when applied to a workspace $F$ as
 only a finite number of accessible terms are present in a given workspace)
 \begin{equation}\label{MergeK}
\cK=\sum_{S,S'} \fM_{S,S'} =\sqcup \circ (\fB\otimes {\rm id}) \circ \Pi_{(2)} \circ \Delta, 
\end{equation}   
where $\Pi_{(2)}$ is the projection that selects the terms with two components in the
left channel of the coproduct. The map \eqref{MergeK} is a Hopf algebra Markov chain.
    \end{itemize}
    
 We refer the reader to \cite{marcolli_mathematical_2024} for a more detailed account
 of this model of syntax and the Merge action. We recalled it here for comparison, to
 easily outline the similarities and differences with the case of morphology and because
 it will become a part of the overall morphosyntactic system.
 
 \section{Modeling Morphology}
 
 As mentioned in the Introduction, our linguistic approach to the mathematical
 modeling of morphology relies on the ideas of Distributed Morphology, as developed
 by Halle and Marantz in \cite{halle_distributed_1993} and \cite{halle_key_1994}. This being said,
 our formulation can be seen as being also, in some respects, related to the
 Nanosyntax approach, in the sense that the basic structure building magma operation is common to both syntax and morphology.
 
 In DM, features are housed in terminal nodes of the syntactic trees. There are rules which 
 manipulate these features on trees: fusion (combining two seperate feature bundles into a 
 singular feature bundle), fission (separating one feature bundle into two distinct ones), impoverishment (removal of one or more features in the feature bundle), 
 and obliteration (the complete elimination of a feature bundle). We will be discussing
 these operations explicitly in \S \ref{DMopsSec} below.
 
 These features are mapped to phonological forms via vocabulary insertion rules which 
 specify individual mappings of feature bundles to a phonological form. Finally, phonology occurs, 
 resulting in the surface phonological forms from the phonological forms specified from the 
 feature-to-phonology mapping, which also takes into account other factors (such as, for instance, vowel harmony).
 In our model phonology is incorporated at Externalization. We focus here only on the
 pre-Externalization part, hence the morphological objects and workspaces that we consider will
 have \textit{only} morphological features at the leaves, not morphemes. This is because our 
 morphological objects and workspaces are language-independent---vocabulary items are only 
 inserted in Externalization, after the morphosyntactic trees have been assembled and 
 reach their final configuration. Note that features can also be considered to be language-dependent, as one can claim that certain features are present in some languages and not in others. However, our morphosyntactic workspace contains the set of all morphological features---if certain features are considered to be incompatible with a given language, these morphological structures/morphosyntactic trees containing those features will be properly filtered out at the Externalization stage.
 
The DM formulation of morphology is very suitable for this approach as it relies on the following two guiding principles: (a) \textit{syntax all the way down}, and (b) \textit{late insertion} (no phonology takes place until the structures are completely built). These ideas very much parallel those of \cite{marcolli_mathematical_2024}, which takes the perspective that the machinery utilized in syntax can also be used in other places, hence minimizing the amount of computational architecture needed by reusing compositional tools already established, e.g. the utilization of syntactic tools (Merge and the coproduct) in the syntax-semantics interface. This philosophy clearly parallels (a). Secondly, with respect to (b), the mathematical model of \cite{marcolli_mathematical_2024} operates with a discrete, sequential approach, where all of the language-independent 
syntax is developed prior to the process of Externalization, and there is a clear division between 
the core computational process of syntax and the interfaces (Externalization at the Sensory-Motor interface 
and the Syntax-Sematics or Conceptual-Intentional interface). The modeling of morphology should be compatible
with this overall organization of the Faculty-of-Language system. 
 
\subsection{Feature bundles and the magma of morphological objects}
 
 The main aspect of our formalization of morphology consists of replacing feature bundles as sets with
 feature bundles as tree structures. We do this in two successive steps. The first step consists of forming
 tree structures (our {\em morphological objects}) by combining features in a hierarchical way, obtaining
 non-planar binary rooted trees with leaves labelled by features and internal vertices labelled by feature bundles.
 
 These morphological objects form a free non-associative commutative magma $\cM\cO$ generated 
 by the set $\cM\cO_0$ of morphological features. All trees obtained in this way are full binary trees. We 
 will then extend this construction in \S \ref{NoBranchSec} so as to obtain also binary trees that contain
 non-branching vertices, as these are of use in representing bundles of features resulting from processes in the DM model of
 morphology. The tree structure of morphological objects is consistent with the idea of feature hierarchies, as formulated for instance in \cite{noyer_1997}. 
 
 Thus, our main claim here is that we can consider bundles of features to have an inherent structure, and we can hence model a feature bundle as tree structures. The idea that features can have hierarchical relations with respect to each other has been explored through the notion of Feature Geometries \citep{harleyPersonNumber2002}, and modeling morphology as tree structures has been proposed at various points to varying degrees. Within nanosyntax, this is a feature of the system, given that the morphological structures are assembled via the syntax \citep{baunaz_nanosyntax_2018,baunaz_exploring_2018}. Moreover, \cite{cowper_syntactic_2002} and \cite{cowper_geometry_2005} utilize feature geometries to inform tree representations of English nominal inflection morphemes and tense-aspect-mood (TAM) morphemes in English and Spanish, respectively. Within DM, \cite{harbour_discontinuous_2008} uses unary-branching trees as the internal structure for phi-features.\footnote{For more instances of representing morphological structures as trees, see \cite{martinovic_feature_2022,svenonius_case_2006,tosco_feature-geometry_2007,coon_feature_2021,martinovic_feature_2022,mcginnis_phi-feature_2007,xu_double-person_2018}, among others.} Generalizing these ideas to make them consistent throughout all morphological features/structures, we take the view that morphological structures (feature bundles) can be modeled as hierarchical tree structures comprised of individual features at leaves that have been iteratively merged. This entails the existence of a morphism between a bundle of features and a binary tree structure. 

 In particular, in the model we develop here, we will not consider morphological trees with higher valence nodes. This is consistent with typical use in Distributed Morphology and with Nanosyntax. However, trees with higher valence nodes are in use in Feature Geometry. Incorporating more general forms of morphological trees is possible. It requires extending the magma we discuss here that generates our morphological tree with additional n-ary operations for higher $n\geq 3$. The use of binary trees simplifies the structure and provides greater consistency in the interface with syntax, as we will be discussing in the following sections. Thus, unless a compelling reason exists for requiring higher valences, the binary structures appear preferable. We will discuss briefly at the end of the paper what changes in the interface with syntax if higher valence morphological trees are used.  

\smallskip

 We introduce, as the basic algebraic structure for morphology, the same fundamental structure we have in the case of syntax, namely a magma operation.
 
 \begin{defn}\label{morphobj}
 Let $\cM\cO_0$ denote the (finite) set of morphological features (such as {\em [$\pm$PL]} for the valued forms
  and {\em [uPL]} for
 the unvalued form of the plural feature, etc.). Let $\cM\cO$ denote the free non-associative commutative magma
 generated by the set $\cM\cO_0$
 \begin{equation}\label{Mtrees}
 \cM\cO ={\rm Magma}_{na,c}(\cM\cO_0, \fM^{\rm morph}) \, .
 \end{equation}
 As in the case of syntactic objects we can identify $\cM\cO \simeq \fT_{\cM\cO_0}$ with the set of non-planar
 full binary rooted trees with leaves decorated by elements of ${\cM\cO_0}$. Unlike in syntax, here we also endow trees
 $S \in \fT_{\cM\cO_0}$ with a labeling of the non-leaf vertices that is completely determined by the labels at the leaves and
 the tree structure. The non-leaf vertices are labelled by sets in $\cP({\cM\cO_0})=2^{{\cM\cO_0}}$ (the power set of ${\cM\cO_0}$) in such a way that, if the two vertices $v_1, v_2$ below a given vertex $v$ are labelled by subsets $B_{v_1}$ and
 $B_{v_2}$ of ${\cM\cO_0}$, then $v$ is labelled by 
 \begin{equation}\label{BvBv12}
 B_v= B_{v_1}\cup B_{v_2}. 
 \end{equation}
 Thus, the root vertex of $S \in \fT_{\cM\cO_0}$ is labelled by the set $\cup_{\ell \in L(S)} B_\ell$, where $L(S)$ is the set of leaves of $M$ and
 $B_\ell=\{ \mu_\ell \}$ is a single feature $\mu_\ell\in \cM\cO_0$ associated to the leaf. We refer to the elements of $\cM\cO \simeq \fT_{\cM\cO_0}$ with this labeling as {\em morphological objects} or {\em morphological trees}, and to the sets 
 $B\in \cP({\cM\cO_0})=2^{{\cM\cO_0}}$ as {\em bundles of features} or {\em feature bundles}.
 \end{defn}

 The labeling of internal vertices of morphological trees shows that we can regard them
 as assembly procedures for bundles of morphological features. These tree structures 
 should be thought of as templates for word formation when morphemes and vocabulary items
 are inserted in Externalization.

 \begin{rem}\label{unionBv}{\rm
 Note that we take $B_v= B_{v_1}\cup B_{v_2}$ in \eqref{BvBv12} rather than $B_{v_1}\sqcup B_{v_2}$.
 This means that, for example, a feature bundle $B_v=[\alpha,\beta,\phi]$ may be obtainable both as
 \begin{equation}\label{Bv2way}
 \Tree[ .$[\alpha,\beta,\phi]$  $\alpha$ [ .{$[\beta, \phi]$} $\beta$ $\phi$ ] ]  \ \ \ \ \text{ or } \ \ \ \  \Tree[ .$[\alpha,\beta,\phi]$  [  .{$[\alpha, \phi]$} $\alpha$ $\phi$ ]  [ .{$[\beta, \phi]$} $\beta$ $\phi$ ] ] 
 \end{equation}
 as well as other possible tree configurations. This fact will be useful in \S \ref{FissionSec}
 to model Distributed Morphology operations like ``fission". Note that free structure
 formation (in morphology as in syntax) typically overgenerates, and that other filtering
 mechanisms intervene to eliminate ill formed structures and tame the combinatorial
 explosion, such as coloring rules (like those governing theta roles and phases in syntax)
 and filtering by morphosyntactic parameters in Externalization. 
  }\end{rem}
  
 The first tree in \eqref{Bv2way}, and in general similar comb-like trees, represent bundles of features that belong to a single feature hierarchy. Feature hierarchies should be modeled by a partial order or a preorder. Thus, other tree topologies will occur to reflect the possibilities of features belonging to different unrelated hierarchies.

 \begin{rem} {\rm
We represent features as bivalent in this work. That being said, the mathematical formulation of morphosyntax developed in this paper is compatible with the choice of either privative (the feature simply exists or does not, so e.g. the plural feature would be represented as [PL] and singular would be represented with a different feature [SG]) or bivalent (every feature can be + or -, so that plural and singular number for example can be represented by [+PL] and [-PL], respectively) features. 

One mathematical argument in favor of features being bivalent is that the size of the feature space is greatly decreased when utilizing the binary(/ternary, including $u$ for unvalued) scale in combination with the feature categories. That is, given $n$ feature categories (number, person, etc.) and three valuations (+,- and $u$), for a total of $n+3$ objects, the combinations can yield $3n$ unique features/feature valuations. On the other hand, this same number of feature valuations would require the much larger (whenever $n > 1$) set of $3n$ objects, as every feature valuation would be expressed as a different feature (plural and singular are different features instead of different valuations of the same PL feature).
}
 \end{rem}

 We also make another assumption about the set of feature
 bundles and the set of lexical items and syntactic features that labels the leaves of the
 syntactic trees.

 In addition to morphological features, hierarchies, and their assembly into morphological trees via the magma operation, we need a rule establishing the ways in which morphological feature bundles can be matched to syntactic data (lexical items and syntactic features in $\cS\cO_0$ at the leaves of syntactic trees).

 \begin{defn}\label{featurematch}
 There is a correspondence through which feature bundles can be matched with lexical items and syntactic features, namely
 a subset $\Gamma_{SM} \subset \cP({\cM\cO_0})\times \cS\cO_0$ 
 such that the second projection $\pi_2:  \cP({\cM\cO_0})\times \cS\cO_0 \to \cS\cO_0$
 restricted to $\Gamma_{SM}$ is surjective
 $$ \pi_2 |_{\Gamma_{SM}} : \Gamma_{SM} \twoheadrightarrow \cS\cO_0 \, . $$
 We refer to $\Gamma_{SM}$ as the {\em Syntax-Morphology feature correspondence}.
 We say that a pair $(B,\alpha)$ consisting of a feature bundle $B\in \cP({\cM\cO_0})$ and an
 element $\alpha \in \cS\cO_0$ is a {\em matching pair} iff $(B,\alpha)\in \Gamma_{SM}$. 
 \end{defn}
 
 The purpose of the correspondence $\Gamma_{SM}$ is to match morphological feature
 bundles to syntactic features and lexical items. It is too restrictive to implement this
 matching through a function, because it may be multivalued (the same bundle of morphological
 features may be compatible with more than one element in $\cS\cO_0$, but also the same
 lexical item may occur with different combinations of morphological features). We do want,
 however, the surjectivity of $\pi_2 |_{\Gamma_{SM}} : \Gamma_{SM} \twoheadrightarrow \cS\cO_0$
 since we want all elements of $\cS\cO_0$ to be able to carry some morphological features. (However, see Remark~\ref{GammaSMnosurj} for a situation where it is preferable to drop this surjectivity condition.)
 
\smallskip

For example, a plural feature in its valued forms [$\pm$PL] can match accompanying nouns 
so that a pair $( [\pm {\rm PL}],\mathtt{N})\in \Gamma_{SM}$. It also exists in its unvalued form [\textit{u}PL],
to be used in syntactic heads like $\mathtt{T}$, with $([{\rm uPL}],\mathtt{T})\in \Gamma_{SM}$,
whose feature is unvalued exactly until  
morphological elements have been plugged into the leaves of the syntax trees to 
create morphosyntactic trees (as we will discuss in \S \ref{MorphoSyntaxSec}).
Later in the system, Agree will target unvalued morphemes such as this unvalued PL morpheme in 
a $\mathtt{T}$-head, because its value is context-sensitive: it will depend on other components of 
the syntax tree (e.g. the subject) which cannot be evaluated at this earlier point in the derivation, 
within the morphological workspace before the morphosyntax trees have been constructed. (This is similar to the coloring
problems for theta roles and phases analyzed in \cite{marcolli_theta_2025} and \cite{marcolli_phases_2025}.)
The formalization of Agree will be discussed separately from this work.

 \subsection{Non-branching vertices and feature bundles}\label{NoBranchSec}

In morphological tree, unlike the case of syntactic trees, it is desirable to also allow non-branching vertices. We can think of trees such as
$$ 
\Tree[.{$[\alpha,\beta,\phi]$} $\alpha$ [ .{$[\beta, \phi]$} $\beta$ ] ]
$$
as representing, in the non-branching node labelled $[\beta, \phi]$,
the addition of a feature $\phi$ that eventually modifies the realization of the feature $\beta$ but does not itself carry a place to be realized as an independent morpheme in vocabulary insertion, unlike the case of a tree of the form
$$ 
\Tree[.{$[\alpha,\beta,\phi]$} $\alpha$ [ .{$[\beta, \phi]$} $\beta$ $\phi$ ] ]
$$
While full binary trees (with no non-branching vertices) can be generated by the magma, incorporating trees with non-branching vertices requires more than just the magma structure. Indeed, this is where morphology makes use of a coproduct structure. We show here that trees with non-braching vertices can be obtained from the morphological trees generated by the magma through the algebraic structure of {\em comodule over a coalgebra}.

 In the modeling of syntax, the coproduct structure on the span of syntactic workspaces
 comes in three different flavors, denoted by $\Delta^c$, $\Delta^\rho$, and $\Delta^d$ in
 \cite{marcolli_mathematical_2024}, with somewhat different algebraic properties 
 (see \S 1.2 of \cite{marcolli_mathematical_2024}). In all of these forms, the left-channel
 of the coproduct is the same, and it contains forests of extracted accessible terms 
 $F_{\underline{v}}=T_{v_1}\sqcup \cdots \sqcup T_{v_k}$, while the difference is in
 the way the remaining term of the extraction, in the right-channel of the coproduct, is
 obtained. In the coproduct $\Delta^c$ the quotients $T/^c F_{\underline{v}}$ are
 obtained by shrinking each accessible term $T_{v_i}$ of $F_{\underline{v}}$ to its
 root vertex $v_i$ that remains labelled by what will be the trace of movement.
 In the coproduct $\Delta^d$ the terms $T/^d F_{\underline{v}}$ are the maximal
 full binary tree obtained by edge contractions from the tree with non-branching
 vertices resulting form cutting off $F_{\underline{v}}$. It is argued in \cite{marcolli_mathematical_2024}
 that these two forms of the coproduct serve different purposes: the one that keeps
 the trace needed for interpretation at the syntax-semantics interface and the
 one that does not keep the trace representing the form at Externalization (where the
 trace is not externalized). In the third coproduct $\Delta^\rho$, intermediate between these two,
 the terms $T/^\rho F_{\underline{v}}$ are non-full binary trees that contain non-branching
 vertices (where the cuts removing $F_{\underline{v}}$ are performed). This form of the
 coproduct does not directly play a role in the model of syntax and the sensory-motor
 (Externalization) and conceptual-intensional (syntax-semantics) interfaces. 
 
 We argue here that, instead, the form $\Delta^\rho$ of the coproduct is useful for
modeling  the interface between syntax and morphology. 
  To this purpose, we first review more carefully the properties of this coproduct.
  
 \subsection{Coproduct and comodule structure} \label{CoprodComodSec}
  
A left comodule $\cN$ for a coalgebra $(\cC,\Delta)$ is a vector space with a linear map
$\rho_L \in \Hom(\cN, \cC\otimes \cN)$ satisfying
$$( {\rm id}_{\cC}\otimes \rho_L) \circ \rho_L = (\Delta \otimes {\rm id}_{\cN}) \circ \rho_L  \ \ \ \text{ and } \ \ \ 
(\epsilon \otimes {\rm id}_{\cN}) \circ \rho_L = {\rm id}_{\cN}\, , $$
with $\epsilon$ the coproduct and counit of $\cC$. A right comodule 
is defined similarly with a linear map $\rho_R \in \Hom(\cN, \cN\otimes \cC)$ staisfying
$$ ( \rho_R \otimes {\rm id}_{\cC}) \circ \rho_R = ({\rm id}_{\cN} \otimes  \Delta) \circ \rho_R  \ \ \ \text{ and } \ \ \ 
({\rm id}_{\cN} \otimes  \epsilon) \circ \rho_R = {\rm id}_{\cN}\, . $$
A bicomodule $\cN$ for a coalgebra $\cC$ is both a left and a right
comodule, with
the compatibility condition expressed by the identity 
\begin{equation}\label{compatcomod}
( {\rm id}_C\otimes \rho_R) \circ \rho_L =(\rho_L \otimes {\rm id}_C) \circ \rho_R\, .
\end{equation}
A coalgebra $\cC$ is a bicomodule over itself with $\rho_L=\rho_R=\Delta$.
  
  \begin{defn}\label{MOext}
  As in \cite{marcolli_mathematical_2024}, we use the notation $\fT^{\leq 2}_{\cM\cO_0}$
  and $\fF^{\leq 2}_{\cM\cO_0}$ for the set of binary rooted trees (respectively, forests) that
  can contain non-branching vertices (so all vertices have $\leq 2$ descendents), with leaves
  decorated by elements of the set $\cM\cO_0$. We denote by $\cV(\fF^{\leq 2}_{\cM\cO_0})$ 
  the vector space (over $\Q$ or
  $\R$) spanned by the forests in $\fF^{\leq 2}_{\cM\cO_0}$. We also write $\cV(\fF_{\cM\cO_0})$
  for the vector space spanned by the forests in the set $\cV(\fF_{\cM\cO_0})$, where
  components are in the set $\cM\cO$ of \eqref{Mtrees}.
  \end{defn}
  
  \begin{rem}\label{bicomodrem}{\rm
  As shown in \cite{marcolli_mathematical_2024}, the vector space $\cV(\fF^{\leq 2}_{\cM\cO_0})$
  is a graded connected Hopf algebra with product and coproduct 
  $(\cV(\fF^{\leq 2}_{\cM\cO_0}), \sqcup, \Delta^\rho)$. The vector space $\cV(\fF_{\cM\cO_0})$
  is a bicomodule over the Hopf algebra $(\cV(\fF^{\leq 2}_{\cM\cO_0}), \sqcup, \Delta^\rho)$. 
  The right-comodule structure is given by 
  \begin{equation}\label{leftcomod}
   \rho_R=\Delta^\rho: \cV(\cF_{\cM\cO_0}) \to \cV(\cF_{\cM\cO_0}) \otimes \cV(\tilde\cF_{\cM\cO_0}^{\leq 2})\, . 
  \end{equation} 
  The unit $1$ of the algebra $(\cV(\cF_{\cM\cO_0}),\sqcup)$ (the formal empty forest) is also the unit of the algebra
$(\cV(\tilde\cF_{\cM\cO_0}^{\leq 2}), \sqcup)$, hence we obtain a left-comodule structure on
$\cV(\cF_{\cM\cO_0})$ by taking $\rho_L = 1 \otimes {\rm id}$. These satisfy the compatibility \eqref{compatcomod}.
  }\end{rem} 
  
 The right-comodule structure is the interesting part here, because it describes the property that, 
 when applying the coproduct $\Delta^\rho$
 to an object $T\in \fT_{\cM\cO_0}$ (or a workspace $F\in \fF_{\cM\cO_0}$) the left-channel of the
 coproduct $\Delta^\rho$ is always also in $\fF_{\cM\cO_0}$, while the right channel is in the larger
 $\fF^{\leq 2}_{\cM\cO_0}$. In other words, because the coproduct is always applying to objects of the morphological magma, all the subtrees of the morphological tree that the coproduct applies to are also elements of that magma, and hence the coproduct can only remove binary-branching morphological trees. On the other hand, the quotient part of the coproduct can leave behind a unary-branching structure that is part of the extended morphological workspace which will be defined in the next subsection.
 
  \subsection{The space of morphological workspaces} \label{MorphoWorkSec}
  
 A significant difference with respect to the syntactic objects is that the morphological
 objects in $\cM\cO$ carry a labeling of the internal vertices by bundles of features 
 $B_v \in \cP(\cM\cO_0)$ constructed on the basis of the features in $\cM\cO_0$ assigned 
 at the leaves. 
 
 \smallskip
 
 The quotients $T/^\rho T_v$ of morphological objects $T\in \cM\cO$, which
 occur in the right-channel of the coproduct $\Delta^\rho$ maintain at the
 non-branching vertices the features that were contributed by the leaves of
 $T_v$. We can see this in the following example.
 
 \begin{ex}\label{quotBex}{\rm
 Consider a morphological tree in $\cM\cO$ of the form
 $$   T=
 \begin{forest}
    [{[$ \alpha, \beta, \gamma,\delta$]} [$\alpha$] [{[$ \beta, \gamma, \delta$]} [$\beta$] [{[$\gamma, \delta$]}[$\gamma$] [$\delta$]]] ]
\end{forest} 
$$
and consider the term in the coproduct $\Delta^\rho(T)$ of the form $T_v \otimes T/^\rho T_v$ where
$T_v=\beta$ is the single leaf marked by the feature $\beta\in \cM\cO_0$. The corresponding quotient
term $T/^\rho T_v$ is of the form
$$ T/^\rho T_v = \Tree[ .{$[ \alpha, \beta, \gamma,\delta] $}  $\alpha$  [ .{$ [ \beta, \gamma, \delta ] $} [ .{$[\gamma, \delta]$}  $\gamma$  $\delta$  ] ] ] $$
} \end{ex}

\smallskip
 
 Note that extended morphological objects also include, for example, single nodes labelled by a bundle of morphological
 features instead of a single feature. These, like the quotient $T/^\rho T_v$  of the previous example, are extended
 morphological trees that cannot be generated by the magma $(\cM\cO,\fM^{\rm morph})$, but they can be generated, starting
 from objects of the magma $\cM\cO$ by applying the coproduct $\Delta^\rho$ and considering the terms that arise in
 the right-channel of the coproduct (that is, they are generated by the right-comodule structure of $\cV(\fF_{\cM\cO_0})$,
 
\smallskip
 
 Thus, in the case of objects in $\fT^{\leq 2}_{\cM\cO_0}$, if we want to keep
 track of the assignments of feature bundles $B_v$ at the internal vertices, in such a
 way that the right-comodule structure still works, we need to allow for more
 general assignments than those determined by the features at the leaves.
 
  \begin{defn}\label{MOext2}
  The {\em extended morphological objects} are pairs $(T,B)$ of a tree $T\in \fT^{\leq 2}_{\cM\cO_0}$
  and an assignment $B: V^o(T) \to \cP(\cM\cO_0)$ with the properties: 
  \begin{itemize}
\item  For $v,w\in  V^o(T)$ with $T_w\subset T_v$ the feature bundles satisfy $B_w\subseteq B_v$.
\item  For all $v\in V^o(T)$,
  the feature bundle $B_v\in \cP(\cM\cO_0)$ contains the set $\cup_{\ell\in L(T_v)} \{ \mu_\ell \}$ of
  all the morphological features $\mu_\ell \in \cM\cO_0$  assigned to the leaves $\ell\in L(T_v)$. 
\item If $T_v\subseteq T$ does not contain any non-branching vertices, then $B_v=\cup_{\ell\in L(T_v)} \{ \mu_\ell \}$,
with $\mu_\ell \in   \cM\cO_0$ the morphological features at the leaves. 
  \end{itemize}
  The {\em morphological workspaces} are forests $F=\sqcup_a T_a$
  where all the components $T_a$ are extended morphological objects.
  We use the notation 
  \begin{equation}\label{morphOW}
  \widetilde{\cM\cO}:=\fT^{\leq 2}_{\cM\cO_0,\fB} \ \ \ \text{ and } \ \ \  \cW_\cM:=\fF(\widetilde{\cM\cO})
  \end{equation}
  for the set $\widetilde{\cM\cO}$ of extended morphological objects, where $\fT^{\leq 2}_{\cM\cO_0,\fB}$
  means the set of pairs $(T,B)$ as above, and the set $\cW_\cM$ of
  morphological workspaces, where $\fF(\widetilde{\cM\cO})$ means the set of forests whose
  components are trees in $\fT^{\leq 2}_{\cM\cO_0,\fB}$. We write $\cV(\cW_\cM)$ for the vector space spanned by the set
  of morphological workspaces. 
  \end{defn}
   
   Thus, the difference between the morphological trees and the extended
   morphological trees is that the latter have non-branching vertices and
   the bundles of features labeling internal vertices can contain
   additional features that are not contained in the set of features assigned
   at the leaves. We will discuss in \S \ref{DMopsSec} why this is
   necessary to properly represent Distributed Morphology.
   
   \smallskip
  
 The assignment of feature bundles $B_v$ to extended morphological objects,
 with the rules of Definition~\ref{MOext2} ensures that the following holds.
 
 \begin{lem}\label{MOtotildeMO}
 The vector space $\cV(\cW_\cM)$ of morphological workspaces is a graded connected
 Hopf algebra with product $\sqcup$ and coproduct $\Delta^\rho$ and the 
 vector subspace $\cV(\fF_{\cM\cO_0})\subset \cV(\cW_\cM)$ is a right-comodule 
 as in \eqref{leftcomod}. 
 \end{lem}
 
 We refer to $(\cV(\cW_\cM),\sqcup,\Delta^\rho)$ as the Hopf algebra
 of morphological workspaces. 
  
 \smallskip
 
 In syntax, the coproduct structure is needed for the extraction of accessible terms (and corresponding cancellation of deeper copies) that are needed for movement. External Merge, by itself, could otherwise be accounted for already by the magma structure of syntactic objects. Workspaces and the coproduct make it possible to unify External Merge and Internal Merge into a single operation definable in Hopf algebra terms. In morphology one does not need movement, so in principle the magma operation would suffice for structure building, except for allowing for non-branching vertices, which as we discussed, can be obtained using the comodule $\rho_R=\Delta^\rho$. Note however that, even though we do use a coproduct/comodule structure, we do not need to introduce in morphology a Merge-like operation: any morphological tree that looks like $\fM(T,T')$ where $T,T'$ are either in $\cM\cO$ or in $\widetilde{\cM\cO}$, is already present in either the magma $\cM\cO$ or in the terms in the right-channel of the coproduct $\Delta^\rho$ applied to elements of the magma $\cM\cO$. 

 Thus, unlike the case of syntax, we do not need to introduce a Merge action on 
 morphological workspaces, rather we will have a different kind of structure
 formation operation, which still relies on the Hopf algebra
 structure, and takes care of interfacing syntax with morphology, leading to
 the creation of morphosyntactic objects. We describe this
 mechanism in the next sections \S \ref{MorphoSyntaxSec} and \S \ref{MorphoSyntaxSec2}.

 The points of view presented in \S \ref{MorphoSyntaxSec} and \S \ref{MorphoSyntaxSec2} describe the formation of morphosyntax 
 from syntactic and morphological data with two slightly 
 different perspectives. In \S \ref{MorphoSyntaxSec} we
 focus on the morphological data inserted at the leaves
 of syntactic trees, hence the algebraic formalism revolves
 around operads and algebras over operads. In \S \ref{MorphoSyntaxSec2} we focus on morphological
 workspaces and operations that use syntactic objects and
 that collect inputs from morphological workspaces to
 produce morphosyntax. These two viewpoints are equivalent
 in terms of the resulting structure formation. It is useful to
 develop both perspective for the same reason discussed in 
 \cite{marcolli_theta_2025} and \cite{marcolli_phases_2025}:
 in view of developing a model of Agreement, it is important
 to be able to formulate filtering of structures via
 the formalism of (colored) operads and algebras over operads,
 while this also need to be implementable alongside
 structure formation, via a description in terms of
 maps acting on workspaces. The results of
 \S \ref{MorphoSyntaxSec} and \S \ref{MorphoSyntaxSec2}
 build the necessary theoretical setting.
 
 \section{Morphosyntax and algebras over operads}  \label{MorphoSyntaxSec}

 The main idea, as in various existing models of morphology, is that morphological features exist as structure
 at the leaves of syntactic trees. An analogy from a physicist's perspective would be that
 the data $\alpha \in \cS\cO_0$ of lexical items and syntactic features at the leaves of syntactic
 objects acquire inner structure (further degrees of freedom) when one zooms in at the scale
 of morphology (word formation) rather than at the large scale structure of syntax (sentence
 formation). This means, in terms of mathematical formulation, that we are looking at an
 operation that inserts structure at the leaves of a tree: this naturally suggests that the
 formalism of operads and algebras over operads is the right type of algebraic structure
 to describe this kind of model. However, as we will see, morphological structures
 do not directly form an algebra over an operad themselves, but the syntactic objects,
 with their structure of an algebra over an operad, determine structure formation operations 
 on the morphological workspaces that combine syntactic objects and morphological
 trees forming the morphosyntactic structures. These
 morphosyntactic objects, in turn, form another algebra over the same operad. 
 
 \smallskip
 \subsection{Operads and algebras over operads}\label{OperSec}
  
 The notion of operad describes the compositions of operations with
 multiple inputs and a single output, where the output of one operation
 can serve as input of another one.
 
 \smallskip
 
An operad (in the category of sets) is a collection
$\fO=\{ \fO(n) \}$ of sets $\fO(n)$ whose elements are operations $T \in \fO(n)$ that have $n$
inputs and one output. The algebraic structure governing this collection of sets $\fO$ 
consists of {\em compositions} that relate the operations in the sets $\fO(n)$. These
compositions are usually presented in two different forms, one that saturates all the
inputs of an operation with outputs of other operations, and one where only one
input at a time is filled with one output. These two different formulations of the
operad structure are equivalent (for unital operads). The formulation with simultaneous
saturation of all inputs is presented as compositions of the form
\begin{equation}\label{operadcomp}
 \gamma: \fO(n) \times \fO(k_1)\times \cdots \times \fO(k_n) \to \fO(k_1+\cdots + k_n) 
\end{equation} 
where $\gamma$ takes the single output of an operation in $\fO(k_j)$ and feeds it into
the $j$-th input of an operation in $\fO(n)$. Since this is done for every input of operations
in $\fO(n)$ the result is now an operation that only has the inputs coming from the inputs of 
the operations in $\fO(k_j)$ (a total of $k_1+\cdots + k_n$ inputs), and a single output, 
and the composition rule \eqref{operadcomp}
requires that the operation obtained in this way is in the set $\fO(k_1+\cdots + k_n)$.
The composition operations \eqref{operadcomp} are also requires to satisfy an 
associativity rule, namely
\begin{equation}\label{gammassoc}
\begin{array}{c} \gamma(\gamma(T,T_1,\ldots,T_n);T_{1,1},\ldots,T_{1,m_1},\ldots,T_{n,1},\ldots, T_{n,m_n})= \\  \gamma(T;\gamma(T_1;T_{1,1},\ldots,T_{1,m_1}),\ldots,\gamma(T_n;T_{n,1},\ldots, T_{n,m_n})) \, . \end{array} 
\end{equation}
The second form of the operad composition, with a single output-input match, has composition
rules (operad insertions) of the form 
\begin{equation}\label{insertions}
 \circ_i: \fO(n)\otimes \fO(m) \to \fO(n+m-1),
\end{equation}
satisfying 
$$
 (X\circ_j Y)\circ_i Z = \left\{ \begin{matrix} (X\circ_i Z)\circ_{j+c-1} Y & 1\leq i < j \\
X\circ_j (Y\circ_{i-j+1} Z)& j\leq i < b+j \\
(X\circ_{i-b+1} Z)\circ_j Y & j+b \leq i \leq a+b-1.  \end{matrix} \right. 
$$
for $1\leq j \leq a$ and $b,c\geq 0$, with  
$X\in \fO(a)$, $Y\in \fO(b)$, and $Z\in \fO(c)$.
When the operad is unital, namely if there is an operation (unit) 
${\bf 1}\in \fO(1)$  that satisfies 
$\gamma({\bf 1};T)=T$ and $\gamma(T;{\bf 1},\ldots,{\bf 1})=T$,
the composition operations \eqref{operadcomp} are equivalent to
the insertion operations \eqref{insertions}, with 
the  laws $\gamma$ of \eqref{operadcomp} obtained from the insertions $\circ_i$ of \eqref{insertions} as 
\begin{equation}\label{gammacircs}
 \gamma(X, Y_1,\ldots, Y_n)=(\cdots (X\circ_n Y_n)\circ_{n-1} Y_{n-1}) \cdots \circ_1 Y_1). 
\end{equation}
In addition to these properties, operads can have a symmetric property that makes
composition rules compatible with permulations of the inputs, but we will not be discussing
them here. 

\smallskip

The notion of algebra over an operad describes sets whose elements can
serve as inputs and outputs of the operations in the operad. 

\smallskip

An algebra $A$ over an operad $\fO$ (in the category of sets) is a set $A$ with an
action of the operad $\fO$, namely operations
\begin{equation}\label{algoper}
\gamma_A: \fO(n) \times A^n \to A  
\end{equation}
satisfying
\begin{equation}\label{gammaAgamma}
\begin{array}{c}
 \gamma_A(\gamma(T;T_1,\ldots, T_n); a_{1,1}, \ldots, a_{1,k_1}, \ldots, a_{n,1}, \ldots, a_{n,k_n}) = \\
 \gamma_A(T; \gamma_A(T_1;  a_{1,1}, \ldots, a_{1,k_1}),\ldots, \gamma_A(T_n; a_{n,1}, \ldots, a_{n,k_n})) \, . 
 \end{array}
 \end{equation}

One interprets here $\gamma_A$ as the operation that takes $n$ inputs $a_i$ from the set $A$ (an element $\underline{a}=(a_i)_{i=1}^n$ in the set $A^n$) and inserts them in the inputs of an $n$-ary operation $T\in \fO(n)$, which then
produces as single output another element of the same set $A$. 

The condition \eqref{algoper} means that one can compose 
operations according to the composition rules $\gamma$ in the operad $\fO$ and then apply them to inputs in $A$, 
or one can {\em equivalently} apply the first operations in $\fO$ to inputs in $A$ using the rule $\gamma_A$ and then 
insert the resulting outputs (also in $A$) as inputs to the second operation in $\fO$, again according to the rule $\gamma_A$. 

We also recall the notion of a colored operad, which
is a collection $\fO=\{ \fO(c, c_1,\ldots, c_n) \}$ of sets, with $c,c_i\in \Omega$ for $i=1,\ldots,n$, where $\Omega$ is a (finite) set of color labels. The $c_i$ are the colors
assigned to the inputs of the $n$-ary operations
in $\fO(c, c_1,\ldots, c_n)$, and $c$ is the color 
label assigned to the output. The 
composition is then like the usual operad composition
but with the requirement that colors should match, namely
\begin{equation}\label{colopergamma}
\begin{array}{rl}
\gamma: & \fO(c, c_1,\ldots, c_n) \times \fO(c_1, c_{1,1}, \ldots, c_{1,k_1}) \times \cdots \times
\fO(c_n, c_{n,1}, \ldots, c_{n,k_n})  \\ & \to \fO(c, c_{1,1}, \ldots, c_{1,k_1}, \ldots, c_{n,1}, \ldots, c_{n,k_n}) \, . \end{array}
\end{equation}
These composition operations satisfy the same 
associativity and unit conditions as in the usual
ones of the non-colored case. Instead of a single 
unit there if now a unit ${\bf 1}_c\in \cO(c,c)$
for each color $c\in \Omega$.

An algebra over a colored operad is 
a collection of sets $A=\{ A^{(c)} \}_{c\in \Omega}$
with the operad action (satisfying the same
compatibility condition with the compositions $\gamma$ as in the non-colored case) of the form
$$ \gamma_A: \fO(c, c_1,\ldots, c_n) \times A^{(c_1)}\times \cdots \times A^{(c_n)} \to A^{(c)} \, . $$

 \subsection{Syntactic objects as an algebra over an operad}  \label{SOoperSec}
 
 Before we can discuss how syntactic objects and morphological objects interface it
 is helpful to recall some results from \cite{marcolli_mathematical_2024} (see also
 \cite{marcolli_theta_2025} and \cite{marcolli_phases_2025}) about realizing
 syntactic objects as an algebra over an operad. 
 
 \smallskip

Operads and algebras over operads have already been used, in modeling 
 syntax, in \cite{marcolli_mathematical_2024}. We recall here how to
 view syntactic objects as an algebra over an operad, as that will be the basis for
 introducing the further structure needed to generate morphosyntactic trees.

\smallskip

The {\em Merge operad} $\cM$ has $\cM(n)$ given by the set of all abstract (non-planar)
binary rooted trees with $n$ (non-labelled) leaves. The insertion operations
$T\circ_\ell T'$, for $\ell\in L(T)$ graft the root vertex of $T'\in \cM(m)$ to
the leaf $\ell$ of $T\in \cM(n)$, resulting in a tree $T\circ_\ell T'$ in $\cM(n+m-1)$.
The set $\cS\cO=\fT_{\cS\cO_0}$ of syntactic objects is then an algebra over the operad $\cM$
with the operad action
$\gamma_{\cS\cO}: \cM(n) \times \fT_{\cS\cO_0}^n \to \fT_{\cS\cO_0}$ that
plugs the root of the syntactic object $T_\ell \in \fT_{\cS\cO_0}$ to the $\ell$-th leaf of the
tree $T\in \cM(n)$. 

\smallskip

We will equivalently use the notation $\gamma(T, T_1,\ldots, T_n)$ or $\gamma(T,\{ T_\ell \}_{\ell\in L(T)})$ for these and other operad insertion operations. With the first notation we do not necessarily mean that the leaves are ordered (the trees are all non-planar), rather that they are labelled so that we know which of the $n$ trees is matched to which leaf, as the second notation clarifies more explicitly. This is discussed also in \S 3.8.1 of 
\cite{marcolli_mathematical_2024}.  

\smallskip

It will be useful in the following (see \S \ref{FissionSec}) to also write the operad
action $\gamma_{\cS\cO}$ as a composition of individual insertions at each of the leaves,
as one does in \eqref{gammacircs} for the operad composition. 
In the case of the operad composition $\gamma$,  the single insertions
\eqref{insertions} map $\circ_\ell : \cM(n) \times \cM(m) \to \cM(m+n-1)$. However, in the
case of operad action $\gamma_{\cS\cO}$, if we perform a single insertion 
of a syntactic object $T'\in \cS\cO$ with $m$ leaves (each of which is labeled by an element of $\mathcal{SO}_0$) at one of the leaves of a $T\in \cM(n)$
the result will be an operation with only $n-1$ inputs, since the $m$ leaves of $T'$ are not open inputs, 
being already filled with elements in $\cS\cO_0$. So we have
\begin{equation}\label{circellSO}
\circ_{\cS\cO_\ell}: \cM(n) \times \cS\cO_m \to \cM_{\cS\cO}(n-1, m)\, ,
\end{equation}
where we write $\cS\cO_m\subset \cS\cO$ for the set of syntactic objects with $m\geq 1$ leaves 
and we use the notation $\cM_{\cS\cO}(n-1, m)$ to indicate the set of non-planar full binary
trees on $n+m-1$ leaves where $n-1$ of the leaves are unlabelled and $m$ are
labelled by elements in $\cS\cO_0$.

The set $\cS\cO$ of syntactic objects 
is not itself an operad, because only the leaves and
not the root of the trees in $\cS\cO$ are labeled,
so elements of $\cS\cO$ can be acted upon by 
elements of $\cM$, but not by other elements of $\cS\cO$.

This leads to an additional remark, which we will not
 discuss in depth in this paper, but that may play a
 useful role in further developments. 

 \begin{rem}\label{colopMO}{\rm
 When elements of $\cS\cO$ are endowed with a head function,
 they have a labeling algorithm which induces 
 a labeling of all the vertices, including
 the root. }
 \end{rem}

Let ${\rm Dom}(h)\subset \cS\cO$ denote
syntactic objects in the domain of a head
function $h: T \mapsto h_T$. (For the definition
and properties of head functions see \S 1.13.3 of
\cite{marcolli_mathematical_2024}.) We write
the elements of ${\rm Dom}(h)$ as pairs
$(T,h_T)$.

\begin{lem}\label{DomhColOp}
The set ${\rm Dom}(h)\subset \cS\cO$ is a colored
operad, with color set $\Omega=\cS\cO_0$.
\end{lem}

\proof As discussed in \S 1.15 of 
\cite{marcolli_mathematical_2024} for a
syntactic object $(T,h_T)$ with a head
function, there is a labeling algorithm
that assigns to each internal (non-leaf)
vertex $v$ of $T$ a label $\alpha_v\in \cS\cO_0$ 
given by the label $\alpha_v=\alpha_{h_T(v)}$ 
at the leaf $h_T(v)$. In particular, the root
vertex $v_0$ of $T$ also carries a label
$\alpha_{v_0}=h_T(v_0)$, the label at the
head of the entire structure $T$. It is
then possible to insert the root
vertex of a syntactic object at a leaf 
vertex of another one as long as their
respective labels match. This gives
${\rm Dom}(h)\subset \cS\cO$ the
structure of a colored operad.
\endproof

A more sophisticated colored operad that also
uses the head function on syntactic objects is
introduced in \cite{marcolli_phases_2025} to
account for the structure of phases.

\begin{rem}\label{OperVsMerge}{\rm
It is important to stress that the operad
structures are not a model of structure formation
(which in the case of syntax is given by the
Merge action on workspaces), rather a 
model for filtering formed syntactic
objects according to coloring conditions
imposed on an underlying non-colored
operad and algebras over this operad,
as in \cite{marcolli_phases_2025}, \cite{marcolli_theta_2025}.
}\end{rem}
 
The operad structure of Lemma~\ref{DomhColOp}
will become relevant when discussing
the interaction of morphology with the coloring
algorithms for syntactic objects that test for
theta roles and phases, as in \cite{marcolli_theta_2025},
\cite{marcolli_phases_2025}, especially in
view of formulating a model for agreement, but we
will not discuss this further in the present paper.
 
 \smallskip
 \subsection{Morphosyntactic trees} \label{MStreeSec} 
 
 The operation of forming morphosyntactic trees consists of operad insertions of
 roots of extended morphological objects $S\in \widetilde{\cM\cO}$ to the leaves of syntactic
 objects $T\in \cS\cO$ with a matching rule $\Gamma_{SM}$ (as in Definition~\ref{featurematch})
 between the feature bundle $B_v\in \cP(\cM\cO)$
 at the root of $S$ and the datum $\alpha\in \cS\cO_0$ at the leaf of $T$ where
 insertion is performed.
 
 \begin{defn}\label{MStreesdef}
 
 We write $\cS\cO_n\subset \cS\cO$ for the subset of syntactic objects with $n$ leaves, and
 $\cS\cO_{\{ \alpha_\ell \}_{\ell \in L}} \subset \cS\cO$ for the subset of syntactic objects $T\in \cS\cO$ with
 set of leaves $L(T)=L$ and with data $\alpha_\ell \in \cS\cO_0$ at the leaves $\ell\in L$, so that
 \begin{equation}\label{SOnL}
  \cS\cO_n =\sqcup_{\# L = n} \cS\cO_{\{ \alpha_\ell \}_{\ell \in L}} \, . 
 \end{equation} 
 Let $\widetilde{\cM\cO}_B\subset  \widetilde{\cM\cO}$ denote the set of extended morphological objects with root vertex
 decorated by the feature bundle $B\in \cP(\cM\cO_0)$.
 The set $\cM\cS$ of {\em morphosyntactic trees} is the range 
 $$ \cM\cS=\bigcup_n  \gamma_{\cS\cO,\cM\cO}(\cS\cO_n \times \widetilde{\cM\cO}^n) $$  
 of the maps
 \begin{equation}\label{gammaMS}
 \gamma_{\cS\cO,\cM\cO}: \cS\cO_n \times \widetilde{\cM\cO}^n \to \cM\cS
 \end{equation}
 with domains 
 \begin{equation}\label{DomMS}
 {\rm Dom}(\gamma_{\cS\cO,\cM\cO}) =\bigcup_n \bigcup_{\# L =n } \left\{ (T, S_1,\ldots, S_n)\in \left(\cS\cO_{\{ \alpha_\ell \}_{\ell \in L}} \times \prod_{\ell\in L} \widetilde{\cM\cO}_{B_\ell}\right) \,|\, (B_\ell, \alpha_\ell) \in \Gamma_{SM} \right\} 
 \end{equation}
 where $\Gamma_{SM}\subset \cP(\cM\cO_0) \times \cS\cO_0$ is the syntax-morphology feature correspondence of
 Definition~\ref{featurematch}. 
 \end{defn}
 
 To illustrate the maps $\gamma_{\cS\cO,\cM\cO}$ of \eqref{gammaMS} consider the following example. 
 
 \begin{ex}\label{exMS}{\rm
 For example, the insertion map $\gamma_{\cS\cO,\cM\cO}: \cS\cO_5 \times \widetilde{\cM\cO}^5 \to \cM\cS$
 performs the insertions of the roots $v_1, \ldots, v_5$ of the extended morphological objects $S_1,\ldots, S_5$
 in the leaves of the syntactic tree $T\in  \cS\cO_5$ as follows:
 \begin{equation}\label{exMStree}
 \begin{forest}for tree={nice empty nodes, s sep = 10}
    [ [,l*=1.45[$\alpha_1$ [$B_{v_1}$,edge = <-  [,l*=.2][,l*=.2 [,l*=.2] [,l*=.2]]] ][$\alpha_2$ [$B_{v_2}$,edge = <- [,l*=.2][,l*=.2] ]]][, l*=.8[$\alpha_3$ [$B_{v_3}$,edge = <- [,l*=.2][,l*=.2 [,l*=.2] [,l*=.2[,l*=.2] [,l*=.2]]]]] [,l*=.5 [$\alpha_4$ [$B_{v_4}$,edge = <- [,l*=.2 [,l*=.2] [,l*=.2]] [,l*=.2 [,l*=.2] [,l*=.2]] ] ] [$\alpha_5$ [$B_{v_5}$,edge = <- [,l*=.2][,l*=.2 [,l*=.2  [,l*=.2] [,l*=.2]]  [,l*=.2]]]]]]]]
\end{forest}
 \end{equation}
 provided that the morphological feature bundles $B_{v_1}, \ldots, B_{v_5}$ at the roots of these morphological trees
 and the lexical items and syntactic features $\alpha_1,\ldots, \alpha_5$ at the leaves of the syntactic tree $T$ satisfy
 the relation
 $(B_{v_\ell}, \alpha_\ell)\in \Gamma_{SM}$ for $\ell=1,\ldots, 5$.
 The morphosyntactic tree \eqref{exMStree} can be written in the form of an insertion 
 $$ \gamma_{\cS\cO,\cM\cO}(T, S_1,\ldots, S_5) $$
with
$$ T= \Tree[ [ $\alpha_1$ $\alpha_2$ ] [ $\alpha_3$ [ $\alpha_4$ $\alpha_5$ ] ] ]  \in \cS\cO $$
and with
$$ S_1 =\Tree[ .$B_{v_1}$  $\phi_{1,1}$ [ $\phi_{1,2}$ $\phi_{1,3}$ ] ]     \ \ \  
S_2 =\Tree[ .$B_{v_2}$   $\phi_{2,1}$  $\phi_{2,2}$  ]   \ \ \  \
S_3 = \Tree[ .$B_{v_3}$  $\phi_{3,1}$ [ $\phi_{3,2}$ [ $\phi_{3,3}$ $\phi_{3,4}$ ] ] ]  $$
$$ S_4 = \Tree[ .$B_{v_4}$ [ $\phi_{4,1}$ $\phi_{4,2}$ ] [  $\phi_{4,3}$ $\phi_{4,4}$ ] ] \ \ \ \ 
S_5= \Tree[ .$B_{v_5}$ $\phi_{5,1}$ [ [ $\phi_{5,2}$ $\phi_{5,3}$ ] $\phi_{5,4}$ ] ]  \ \ \ \in \widetilde{\cM\cO}
$$
 } \end{ex}
 
 \smallskip
 
 Like syntactic objects, the resulting morphosyntactic trees also form an algebra over an operad.
 
 \begin{lem}\label{MSalgop}
 The set $\cM\cS$ is an algebra over the Merge operad  with
 $$ \gamma_{\cM\cS}: \cM(n) \times \cM\cS^n \to \cM\cS $$
 that grafts the root vertices of the $n$ morphosyntactic trees to the unlabelled leaves of the
 trees in $\cM(n)$.
 \end{lem}
 
 \proof The operad action $\gamma_{\cM\cS}$ of the Merge operad $\cM$ on morphosyntactic trees
 is directly induced by the operad action $\gamma_{\cS\cO}$ on syntactic objects, since the root
 vertex of a morphosyntactic tree is the root vertex of a syntactic object. 
 \endproof

\begin{rem}\label{MSh}{\rm
In a similar way, if we consider 
${\rm Dom}(h) \subset \cS\cO$ with
the colored operad structure of 
Lemma~\ref{DomhColOp}, the subset $\cM\cS_h\subset \cM\cS$ of
morphosyntactic trees of the form
$$ \cM\cS_h=\{ \gamma_{\cS\cO,\cM\cO}(T,S_1,\ldots,S_n) \,|\, 
T\in {\rm Dom}(h) \, \text{ and } \, S_i \in 
\widetilde{\cM\cO} \} $$
is an algebra over the colored operad ${\rm Dom}(h)$.}
\end{rem}

\smallskip
\subsection{Morphological trees}\label{MOnoperSec}

We have seen that both $\cS\cO$ and $\cM\cS$ are algebras over the
Merge operad $\cM$. 
There is, instead, an important difference in the case of the
 sets $\cM\cO$ and $\widetilde{\cM\cO}$.
 
 \begin{lem}\label{noAlgOp} 
 The sets $\cM\cO$ and $\widetilde{\cM\cO}$ of morphological and extended morphological trees are not algebras over the Merge operad $\cM$.
 \end{lem}

 \proof Operations in $\cM$ are full binary trees with
 unlabeled leaves. An action of $\cM$ on the set 
 $\cM\cO$ or $\widetilde{\cM\cO}$ would require using
 morphological trees as inputs for elements of $\cM$,
 but the labels $B_v$ at the root of the morphological 
 tree require syntactic information to combine with
 that the syntactic objects in $\cS\cO$ have, in the
 form of elements $\alpha \in \cS\cO_0$ at the leaves,
 but that operations in $\cM$ do not have. We can think
 of this requirement, in mathematical terms as
 a coloring of the root of the morphological tree that
 requires a matching coloring of leaves for an
 insertion to take place, but there is no coloring
 of the leaves of elements $\cM$. 
\endproof

 In the same way, while trees in $\cM\cO$ 
 or $\widetilde{\cM\cO}$ can
 be inserted at the leaves of trees in $\cS\cO$,
 this again does not give these sets the structure of
 algebras over an operad, in the case
 of ${\rm Dom}(h) \subset \cS\cO$ with
 its colored operad structure of Lemma~\ref{DomhColOp}, 
 since the result of such insertions would be
 morphosyntactic and not morphological trees.

 \smallskip

 \subsection{Correspondences of algebras over operads}\label{CorrAlgOpSec}

 We have seen that both the set $\cS\cO$ of syntactic objects and the set $\cM\cS$ of morphosyntactic objects are algebras over the same operad $\cM$. It is then natural to ask what is the relation between these two algebras-over-operads. The question of how to properly formulate this relation is relevant because the structure of algebra over an operad for syntactic objects is important to describe filtering of freely formed structures produced by Merge, especially filtering for theta role assignments, as in \cite{marcolli_theta_2025}, and filtering for well-formed phases, as in \cite{marcolli_phases_2025}. These filtering are formulated in terms of coloring rules on these operads and algebras over operads. Morphosyntactic trees will also have to undergo similar filtering procedures in terms of colored operads, subject to compatibility relations with the corresponding structures on syntax. Most importantly, we expect such compatibilities of colored operads and algebras over operads to play an important role in the modeling of Agreement. Thus, it is important to provide a good formulation of the relation between these two algebras, $\cS\cO$ and $\cM\cS$, over the Merge operad $\cM$. 

 The usual way in which one compares algebras over operads it through the following notion of morphisms.

 \begin{defn}\label{morAO0}
 A morphism $\varphi: A \to B$ of algebras over an operad $\fO$ is a map of sets satisfying the commutative diagram
 \begin{equation}\label{morAOeq}
  \xymatrix{ \fO(n) \times A^n \ar[r]^{\qquad\gamma_A} \ar[d]_{{\rm id}\times \varphi^n} & A \ar[d]^\varphi \\
 \fO(n) \times B^n \ar[r]^{\qquad\gamma_B} & B } 
 \end{equation}
 \end{defn}

There is, in this sense, a relation between $\cS\cO$
and $\cM\cS$, which is simply given by the ``forgetful morphism" 
$\varphi: \cM\cS \to \cS\cO$ from morphosyntactic trees to syntactic trees that forgets the morphology by shrinking the
morphological subtrees $S_\ell$ to their root vertex $\ell\in L(T)$, dropping the $B_\ell$ part of the label $(B_\ell,\alpha_\ell)$ at this vertex, resulting in just the syntactic tree:
$$ \varphi: \gamma_{\cS\cO,\cM\cO}(T, \{ S_\ell \}_{\ell\in L(T)} ) \mapsto T \, . $$
This clearly satisfies \eqref{morAOeq}. 

However, there is another more interesting
relation that follows the process of structure
formation and that involves the insertion of
morphology at the leaves of syntactic trees,
rather than the removal of morphology from
morphosyntactic trees. 

\begin{ex}\label{MS2ways}{\rm For example, 
consider again the case of the morphosyntactic
tree of \eqref{exMStree}. We can obtain it
by first applying the operad action to
syntactic trees and then inserting morphological
trees in the resulting syntactic tree, namely
first forming
$$ \gamma_{\cS\cO,\cM\cO}: \cM(3)\times \cS\cO_2^3 \to \cS\cO\, , \ \ \ \   T = \gamma_{\cS\cO}(T', T'_1,\ldots, T'_3) $$
with
$$ T' = \Tree[  $\bullet$  [ $\bullet$  $\bullet$ ] ]  \in \cM(3) $$
where $\bullet$ marks the an inputs of operations in the operad $\cM$, and
$$ T'_1 = \Tree[ $\alpha_1$ $\alpha_2$ ] \ \ \ \  T'_2 = \alpha_3 \ \ \ \  T_3'  = \Tree[ $\alpha_4$ $\alpha_5$ ] \ \ \  \in \cS\cO $$
and then inserting the morphological trees $S_1,\ldots, S_5$ at the leaves of $T$ with the operation
$$ \gamma_{\cS\cO,\cM\cO}(T, S_1,\ldots, S_5) $$
or by first inserting morphological trees in the
syntactic trees, and then applying the operad
action to the resulting morphosyntactic trees.
This means that we can first insert the morphological trees by forming
$$ \gamma_{\cS\cO,\cM\cO}(T'_1, S_1,S_2)   =\Tree[ [ .$(B_{v_1},\alpha_1)$  $\phi_{1,1}$ [ $\phi_{1,2}$ $\phi_{1,3}$ ] ]  [ .$(B_{v_2},\alpha_2)$   $\phi_{2,1}$  $\phi_{2,2}$  ] ] \ \ \ \ 
 \gamma_{\cS\cO,\cM\cO}(T'_2, S_3)   = \Tree[ .$(B_{v_3},\alpha_3)$  $\phi_{3,1}$ [ $\phi_{3,2}$ [ $\phi_{3,3}$ $\phi_{3,4}$ ] ] ] $$
$$ \gamma_{\cS\cO,\cM\cO}(T'_3, S_4,S_5) = \Tree[ [ .$(B_{v_4},\alpha_4)$ [ $\phi_{4,1}$ $\phi_{4,2}$ ] [  $\phi_{4,3}$ $\phi_{4,4}$ ] ] 
[ .$(B_{v_5},\alpha_5)$ $\phi_{5,1}$ [ [ $\phi_{5,2}$ $\phi_{5,3}$ ] $\phi_{5,4}$ ] ]  ] $$
and then acting with the operad,
$\gamma_{\cS\cO,\cM\cO}: \cM(3)\times \cM\cS_2^3 \to \cM\cS$ to obtain 
$$ \gamma_{\cS\cO,\cM\cO}(T', \gamma_{\cS\cO,\cM\cO}(T'_1, S_1,S_2) , \gamma_{\cS\cO,\cM\cO}(T'_2, S_3) ,\gamma_{\cS\cO,\cM\cO}(T'_3, S_4,S_5) )\, .  $$ }
\end{ex}

\smallskip

Again, as pointed out in Remark~\ref{OperVsMerge}, these
operadic structures are not in themselves 
a model of structure formation: we will come to
that more explicitly in \S \ref{MorphoSyntaxSec2}
and we will discuss how the insertion operation
$\gamma_{\cS\cO,\cM\cO}$ is involved. The analysis
of these algebras over operads and their relation
is discussed here as preliminary to a theory of
filtering of morphosyntactic structures analogous
to the filtering of syntactic structures
described in \cite{marcolli_theta_2025} and
\cite{marcolli_phases_2025}.

\smallskip

\begin{rem}\label{SOMSgraded}{\rm 
Note that, in the case we are considering, not only the operad is graded by the number of inputs, $\cM=\sqcup_{n\geq 1} \cM(n)$, but both the algebras $\cS\cO$ and $\cM\cS$ over this operad also have a grading $\cS\cO=\sqcup_{n\geq 1} \cS\cO_n$ and $\cM\cS=\sqcup_{n\geq 1} \cM\cS_n$, where $\cS\cO_n$ and $\cM\cS_n$ denote the set of syntactic (respectively, morphosyntactic) trees with $n$ leaves.  }
 \end{rem}

 We also have a grading by number of leaves $\widetilde{\cM\cO}=\sqcup_n\widetilde{\cM\cO}_n$ on extended morphological trees. 
However, as discussed in Lemma~\ref{noAlgOp}, this set is not  
an algebra over the Merge operad.

 \smallskip

A first step 
to describe more explicitly the relation between the two algebras $\cS\cO$ and $\cM\cS$ over
 the Merge operad illustrated in Example~\ref{MS2ways}, it is convenient to first refine the notion of algebra over 
 an operad to a graded version, which we define 
 in the following way.

\begin{defn}\label{opergradA}
Let $\fO$ be an operad and $A$ an algebra over an operad, in the category of sets. 
We say that $A$ is a {\em graded algebra over the operad} $\fO$ if 
$A=\sqcup_n A_n$ for $n\geq 1$ and the operad action maps $\gamma_A: \fO(n) \times A^n \to A$
satisfy
\begin{equation}\label{gradedAO}
\gamma_A: \fO(n) \times A_{k_1}\times \cdots \times A_{k_n}\to A_{k_1+\cdots + k_n} \, ,
\end{equation}
so that \eqref{gammaAgamma} takes the form of the commutative diagram 
$$
\xymatrix{ X \ar[dd]^{\sigma} \ar[r]^{\gamma \times {\rm id}}  &   Y \ar[dr]_{\gamma_A} &  \\
& & Z \\ 
U \ar[r]^{{\rm id} \times \gamma_A^n} & V \ar[ur]^{\gamma_A} & }
$$
with $\sigma$ the permutation of the factors and 
$$ X = \fO(n) \times \fO(k_1) \times \cdots \times \fO(k_n) \times A_{\ell_{1,1}} \times \cdots \times A_{\ell_{1,k_1}}\times
\cdots \times A_{\ell_{n,1}} \times \cdots \times A_{\ell_{n,k_n}} \, , $$
$$ Y= \fO (k_1+\cdots + k_n) \times A_{\ell_{1,1}} \times \cdots \times A_{\ell_{n,k_n}} \, , $$
$$ U= \fO(n) \times \fO(k_1) \times A_{\ell_{1,1}} \times \cdots \times A_{\ell_{1,k_1}}\times \cdots
\times \fO(k_n) \times A_{\ell_{n,1}} \times \cdots \times A_{\ell_{n,k_n}} \, , $$
$$ V= \fO(n) \times A_{\ell_{1,1}+\cdots \ell_{1,k_1}} \times \cdots \times A_{\ell_{n,1}+\cdots + \ell_{n,k_n}} \, , $$
$$ Z =A_{\ell_{1,1}+ \cdots + \ell_{n,k_n}} \, .$$
\end{defn}

 Given that both $\cS\cO$ and $\cM\cS$ are
 graded algebras over the operad $\cM$, as in
 Definition~\ref{opergradA}, we would like to
 formulate their relation in a way that this
 graded structure is taken into account.
 
 There is a straightforward way of 
 extending the usual notion of morphism of
 algebras over operads as in Definition~\ref{morAO0} to
 the graded case in the following way.

 \begin{defn}\label{morAO}
 Similarly, a morphism of graded algebras over an operad $\fO$ is a collection of maps $\varphi_n: A_n \to B_n$
 satisfying the commutative diagrams, for all $n\geq 1$
 $$ \xymatrix{ \fO(n) \times A_{k_1} \times \cdots \times A_{k_n} \ar[r]^{\qquad\gamma_A} \ar[d]_{{\rm id}\times \varphi_{k_1}\times \cdots \times \varphi_{k_n}} & A_{k_1+\cdots + k_n} \ar[d]^{\varphi_{k_1+\cdots + k_n}} \\
 \fO(n) \times B_{k_1} \times \cdots \times B_{k_n} \ar[r]^{\qquad\gamma_B} & B_{k_1+\cdots + k_n} } $$
  \end{defn}

  However, it is clear that this
  simple generalization is not what we need.
  Indeed, one can immediately observe, for 
  example, that the forgetful morphism
  $\varphi: \cM\cS \to \cS\cO$ that shrinks
  the morphological trees to their root vertices
  is not a morphism of graded
  algebras over the $\cM$ operad, as it obviously
  does not preserve degrees, since all the leaves of
  each morphological tree are identified to the
  same leaf of the image in $\cS\cO$. 

  Since we are interested not so much in this
  forgetful morphism but rather in the operation
  of {\em adding} morphology to syntactic trees, 
  we formulate a generalization of morphisms
  of graded algebras over operads that will
  account for this transition from syntactic
  to morphosyntactic trees.

  \smallskip
  
  We extend the notion of morphisms of (graded) algebras over operads to a more flexible notion of
  {\em correspondences}. This replaces directly mapping the sets $\varphi_n: A_n \to B_n$, by maps involving
  auxiliary sets, which make it possible to consistently change degrees compatibly with the operad action. 

  \smallskip
  
  \begin{defn}\label{CorrAB}
  A correspondence $\cC=(C, \gamma_{A,C}): A \to B$ between graded algebras $A=\cup_n A_n$
  and $B=\cup_n B_n$ over an operad $\fO$ is a collection of sets $C=\cup_n C_n$ and maps
  \begin{equation}\label{gammaAC}
  \gamma_{A,C} : A_n \times C_{k_1} \times \cdots C_{k_n} \to B_{k_1+\cdots + k_n} 
  \end{equation}
  for all $n, k_1,\ldots, k_n \geq 1$, 
  that satisfy the compatibility properties with the operad actions $\gamma_A$ and $\gamma_B$
  given by commutative diagrams of the form
  $$
\xymatrix{ X \ar[dd]^{\sigma} \ar[r]^{\gamma_A \times {\rm id}}  &   Y \ar[dr]_{\gamma_{A,C}} &  \\
& & Z \\ 
U \ar[r]^{{\rm id} \times \gamma_{A,C}^n} & V \ar[ur]^{\gamma_B} & }
$$
with $\sigma$ the permutation of the factors and 
$$ X = \fO(n) \times A_{k_1} \times \cdots \times A_{k_n} \times C_{\ell_{1,1}} \times \cdots \times C_{\ell_{1,k_1}}\times
\cdots \times C_{\ell_{n,1}} \times \cdots \times C_{\ell_{n,k_n}} \, , $$
$$ Y= A_{k_1+\cdots + k_n} \times C_{\ell_{1,1}} \times \cdots \times C_{\ell_{n,k_n}} \, , $$
$$ U= \fO(n) \times A_{k_1} \times C_{\ell_{1,1}} \times \cdots \times C_{\ell_{1,k_1}}\times \cdots
\times A_{k_n} \times C_{\ell_{n,1}} \times \cdots \times C_{\ell_{n,k_n}} \, , $$
$$ V= \fO(n) \times B_{\ell_{1,1}+\cdots \ell_{1,k_1}} \times \cdots \times B_{\ell_{n,1}+\cdots + \ell_{n,k_n}} \, , $$
$$ Z =B_{\ell_{1,1}+ \cdots + \ell_{n,k_n}} \, .$$
  \end{defn}
  
  \smallskip

  Here we do not require the operad $\fO$ to be a colored operad, although that may be the
case when syntactic objects are filtered for consistent theta roles assignments and for
well formed phases, as in \cite{marcolli_theta_2025} and \cite{marcolli_phases_2025}. 
However, we still do need a {\em colored version} of the correspondences of Definition~\ref{CorrAB}.

\begin{defn}\label{opermodCol}
Suppose given finite sets $\Omega_A$, $\Omega_B$ and $\Omega_C$. 
Assume given a correspondence between these two sets, in the form of a subset $\Gamma \subset \Omega_C \times \Omega$.
Suppose $A$ is a graded algebras over an ordinary (non-colored) operad $\fO$, as in Definition~\ref{opergradA},
with the property that, for all $n\geq 1$, the sets $A_n$ decompose as
$A_n=\sqcup_{a_1,\ldots, a_n \in \Omega_A} A_{a_1,\ldots, a_n}$, 
so that the maps \eqref{gradedAO} restrict to maps
\begin{equation}\label{gradAOcol}
\gamma_A : \fO(n) \times A_{a_{1,1},\ldots, a_{1,k_1}} \times \cdots \times A_{a_{n,1}, \ldots, a_{n,k_n}} \to A_{a_{1,1},\ldots, a_{n,k_n}} \, .
\end{equation}
We say that $A$ is a graded colored algebra over $\fO$.
A {\em colored correspondence} $\cC=(C,\gamma_{A,C})$ of a graded colored algebras $A$ and $B$ over the operad $\fO$
is a collection of sets $C=\sqcup_n C_n$ with 
$$ C_n=\bigsqcup_{u \in \Omega_C \,\, b_1,\ldots, b_n \in \Omega_B} C^u_{b_1,\ldots, b_n} $$ with the property that the
maps $\gamma_{A,C}$ of \eqref{gammaAC} restrict to maps 
\begin{equation}\label{gammaACcol}
\gamma_{A,C}: A_{c_1,\ldots, c_n} \times C^{u_1}_{b_{1,1},\ldots, b_{1,k_1}}\times \cdots \times C^{u_n}_{b_{n,1},\ldots, b_{n,k_n}} \to B_{b_{1,1},\ldots, b_{n,k_n}} 
\end{equation}
defined on the domains
\begin{equation}\label{DomAC}
 {\rm Dom}(\gamma_{A,C})=\{ (x, y_1,\ldots, y_n) \,|\, x\in A_{c_1,\ldots, c_n}, \, \, y_i \in C^{u_i}_{b_{i,1},\ldots, b_{i,k_i}} \,\, \text{ with } (u_i, c_i)\in \Gamma \}.  
\end{equation} 
The maps \eqref{gammaACcol} satisfy compatibility with the maps $\gamma_A$ and $\gamma_B$ of graded colored
algebras over $\fO$, as in \eqref{gradAOcol}, of the same form as the diagrams in Definition~\ref{CorrAB}
with the colored decompositions of the sets $A,B,C$ taken into account.
\end{defn}

\smallskip

\begin{thm}\label{MOcorr}
The set $\widetilde{\cM\cO}$ of extended morphological objects is a colored correspondence 
between the algebras over the Merge operad $\cM$ given by the set $\cS\cO$ of syntactic
objects and the set $\cM\cS$ of morphosyntactic trees, with color sets $\Omega_{\cS\cO}=\cS\cO_0$,
$\Omega_{\cM\cS}=\cP(\cM\cO_0)$, and $\Omega_{\widetilde{\cM\cO}}=\cP(\cM\cO_0)$.
\end{thm}

\proof We just need to check that the maps $\gamma_{\cS\cO}$ that give the
Merge operad action on syntactic objects and $\gamma_{\cM\cS}$, the
operad action on morphosyntactic trees, are compatible with the maps
$\gamma_{\cS\cO,\cM\cO}$ of \eqref{gammaMS} through commutative 
diagrams as in Definition~\ref{CorrAB}, for the colored version as in 
Definition~\ref{opermodCol}. This is the case since the colors $\Omega_{\cM\cS}=\cP(\cM\cO_0)$
and $\Omega_{\widetilde{\cM\cO}}=\cP(\cM\cO_0)$ are the feature bundles (or single features)
at the leaves of both extended morphological objects and morphosyntactic trees and
the colors $\Omega_{\cS\cO}=\cS\cO_0$ are the lexical items at the leaves of the syntactic objects
and the operad insertions of roots of extended morphological objects at leaves of syntactic
trees that forms morphosyntactic trees is constrained by the Syntax-Morphology feature correspondence
of Definition~\ref{featurematch} that gives the domains as in \eqref{DomAC}. The
compatibility of all the maps in the diagrams then follows. 
\endproof

\section{Structure building in Morphosyntax}  \label{MorphoSyntaxSec2}

We now combine the structures we have discussed so far involving morphological objects
and their interfacing with syntactic objects. In particular, we combine the existence of the
coproduct $\Delta^\rho$ on the span of morphological workspaces with the operadic
insertion maps $ \gamma_{\cS\cO,\cM\cO}$ of \eqref{gammaMS} relating morphological
objects, syntactic objects and morphosyntactic trees.

In syntax, Merge, in the form \eqref{MergeSS} or in the assembled form \eqref{MergeK}, 
is the fundamental structure building operation. In the interface between syntax and
morphology, structure building in syntax has been completed, hence the syntactic objects $T\in \cS\cO$
are available as material for the structure building of morphosyntax. Each  syntactic object
$T\in \cS\cO$ contributes a structure building operation $\cK_T$ that takes material from the
morphological data (the morphological workspaces) and assembles corresponding
morphosyntactic trees, using the operadic insertions $\gamma_{\cS\cO,\cM\cO}$
described above. The key property of the operation $\cK_T$ is that it combines the
insertion operations $\gamma_{\cS\cO,\cM\cO}$ with the product $\sqcup$ and coproduct
$\Delta^\rho$ structure of the Hopf algebra of morphological workspaces. 

\begin{defn}\label{KTdef}
Let $\cW_{\cM\cS}=\fF(\cM\cS)$ denote the set of morphosyntactic workspaces,
the set of forests whose components are morphosyntactic trees, and let $\cV(\cW_{\cM\cS})$
be the vector space spanned by these forests. 
Consider the following linear maps on $\cK_T: \cV(\cW_\cM) \to \cV(\cW_{\cM\cS})$, for
$T\in \cS\cO$ a syntactic object with data $\alpha_\ell\in \cS\cO_0$ at the leaves $\ell\in L=L(T)$, 
defined as
\begin{equation}\label{KT}
\cK_T := \sqcup \circ (\gamma_{\cS\cO,\cM\cO}(T, \ldots ) \otimes {\rm id}) \circ \delta_{\underline{B},\underline{\alpha}_L} \circ \Delta^\rho \, ,
\end{equation}
where $\delta^{\underline{B}}{\underline{\alpha}_L}$, for $B=(B_\ell)_{\ell\in L}$ and 
$\underline{\alpha}_L=(\alpha_\ell)_{\ell\in L}$, is the linear operator defined on the basis elements 
as follows and then extended by linearity:
$$ \delta_{\underline{B}, \underline{\alpha}_L} (F_{\underline{v}}\otimes F/^\rho F_{\underline{v}}) = \left\{ \begin{array}{ll} F_{\underline{v}}\otimes F/^\rho F_{\underline{v}} & F_{\underline{v}}=\sqcup_{\ell\in L} T_\ell \, \text{ with } B_{v_\ell}=B_\ell  \text{ and } (B_\ell, \alpha_\ell) \in \Gamma_{SM} \\
0 & \text{otherwise.}
\end{array} \right. $$
where $B_{v_\ell}\in \cP(\cM\cO_0)$ is the bundle of morphological features assigned to the root $v_\ell$ of the 
extended morphological tree $T_\ell$ that will be inserted at the
leaf $\ell \in L=L(T)$ of the syntactic tree $T$.
\end{defn}

The operations $\cK_T$ defined as in \eqref{KT} resemble the form of the Merge operation in \eqref{MergeK}, hence we are using a similar notation.

We can also formulate an analog of the Merge operations $\fM_{S,S'}$ of \eqref{MergeSS}, for which we will also use
a similar notation $\fM^T_{S_1,\ldots, S_n}$ (see \eqref{MTSell} below). These operations
isolate individual terms of $\cK_T$, as in \eqref{KTMTS}, which
is an analog of the sum in \eqref{MergeK}. While formally these operations look similar to their syntactic counterparts, there are important differences:
\begin{itemize}
\item Unlike the syntactic Merge of \eqref{MergeSS} and \eqref{MergeK}, the operations $\cK_T$ and 
$\fM^T_{S_1,\ldots, S_n}$ use already formed 
syntactic and morphological objects to assemble morphosyntactic objects, hence they do not map to the same space, hence they cannot define a dynamical system by iteration (unlike the Hopf algebra Markov chain of Merge, that gives the syntactic derivations).
\item The operations $\cK_T$ and $\fM^T_{S_1,\ldots, S_n}$
are post-syntactic, in the sense that they rely on the products
of syntactic Merge and model the interface of syntax with
morphology.
\end{itemize}

For $T\in \cS\cO_n$
and $S_1,\ldots, S_n\in \widetilde{\cM\cO}$ such that, for all $\ell \in L(T)$ the pair $(B_\ell, \alpha_\ell)$ is in $\Gamma_{SM}$
for $B_\ell$ the morphological feature bundle at the root of $S_\ell$ and $\alpha_\ell \in \cS\cO_0$ at the corresponding
leaf of $T$, we set 
\begin{equation}\label{MTSell}
\fM^T_{S_1,\ldots, S_n}= \sqcup \circ (\gamma_{\cS\cO,\cM\cO}(T, \ldots ) \otimes {\rm id}) \circ \delta_{S_1,\ldots, S_n} \circ \Delta^\rho \, ,
\end{equation}
where the linear operator $\delta_{S_1,\ldots, S_n}$ is defined on the basis elements as follows and extended by linearity:
$$ \delta_{S_1,\ldots, S_n} (F_{\underline{v}}\otimes F/^\rho F_{\underline{v}}) = \left\{ \begin{array}{ll} F_{\underline{v}}\otimes F/^\rho F_{\underline{v}} & F_{\underline{v}}=\sqcup_\ell S_\ell \\ 
0 & \text{otherwise.}
\end{array} \right. $$
These satisfy
\begin{equation}\label{KTMTS}
 \cK_T= \sum_{S_1,\ldots, S_n} \fM^T_{S_1,\ldots, S_n} 
\end{equation} 
where the sum is taken over all $S_1,\ldots, S_\ell$ satisfying the conditions $(B_\ell, \alpha_\ell) \in \Gamma_{SM}$.
While the right-hand side is formally an infinite sum, it is always a finite sum when applied to a given morphological workspace.
These operations allow us to extract terms from the morphological workspaces via $\Delta^\rho$
and insert them at the leaves of syntactic trees, resulting in morphosyntactic trees. 

\begin{defn}\label{LSOMOdef}
We denote by 
\begin{equation}\label{LSOMO}
\cL_{\cS\cO,\cM\cO}=\{ \fM^T_{S_1,\ldots, S_n} \,|\, T\in \cS\cO, \, S_1,\ldots, S_n \in \widetilde{\cM\cO} \} 
\end{equation}
the set of all the linear operators of the form \eqref{MTSell} that interface syntactic objects with
morphological trees building resulting morphosyntactic structures.
\end{defn}

We will see in \S \ref{DMopsSec} certain well known operations of Distributed Morphology can be
seen as transformations acting on this set $\cL_{\cS\cO,\cM\cO}$.

\section{The operations of Distributed Morphology}\label{DMopsSec}

The four DM operations of fission, fusion, impoverishment, and obliteration operate on morphosyntactic trees as post-syntactic operations which manipulate the morphosyntactic tree structures. We now approach the mechanics of the four DM operations. We will present them in two different but equivalent perspectives. First we reformulate in our setting the usual way in which these operations are described in the DM literature, by presenting them as transformations of morphosyntactic trees. It should be pointed out that they are often seen just as transformations of morphological trees (or bundles of features), but in fact the way the morphological structures are inserted at the leaves of syntactic trees matters in definining these operations, as will be clear in the following, so they should be regarded as acting on fully formed morphosyntactic trees. There is, however, another equivalent viewpoint that we will present, that identifies these operations of DM as transformations acting on the set of all the structure-building operations $\{ \fM^T_{S_1,\ldots, S_n} \}$ of
morphosyntax. In this prespective, the DM operations are not so
much altering morphological or morphosyntactic trees, but
rather altering the recipes for assembling morphosyntactic trees. 

\smallskip

As we will see more in detail below, fusion pushes the morphological part of morphosyntax upward into the syntactic part
(by changing a syntactic vertex of the morphosyntactic tree into a morphological vertex), while fission does the opposite operation, pushing syntax downward into the morphology part, transforming a morphological vertex into a syntactic vertex. While these two operations appear in this sense symmetric there is a key difference in their formalization: fusion only uses the magma operation that is common to both morphology and syntax, while fission is a truly morphological operation that also involves set-theoretic operations on bundles of morphological features and could not exist within syntax. This difference is not surprising: since the fusion operation moves toward syntax, it should rely only on operations that are available within syntax, while fission that moves toward morphology relies also on purely morphological data. The remaining two operations, obliteration and impoverishment, rely on the coproduct $\Delta^\rho$ and (in the case of impoverishment) on the fission operation. 

\subsection{Fusion} \label{FusionSec}

Fusion refers to two different morphological feature bundles in two different adjacent 
syntactic leaves (two leaves of a cherry subtree) being merged into one feature bundle. 
This can apply, for instance, to the case of two different adjacent heads that are merged 
together as derived from head-to-head movement (for a discussion of this type of
head-to-head movement in the context of the larger mathematical formulation see \cite{marcolli_extension_2024}).
This fusion mechanism is typically thought to also combine the two leaves together, 
so that the resulting feature bundle is assigned to the single remaining leaf vertex. 
That is, the syntactic tree itself is also modified, with a cherry tree contracted to its
root vertex, which becomes the new leaf.

An explicit example of fusion can be seen in negation in Swahili \cite{nevins2016lectures}.

\begin{ex}{\rm
    In the case of 1PL in Swahili, ``We will love Swahili'' and its negation ``We will not love Swahili'' are stated in (a) and (b), respectively:

    \includegraphics[width = .37\textwidth]{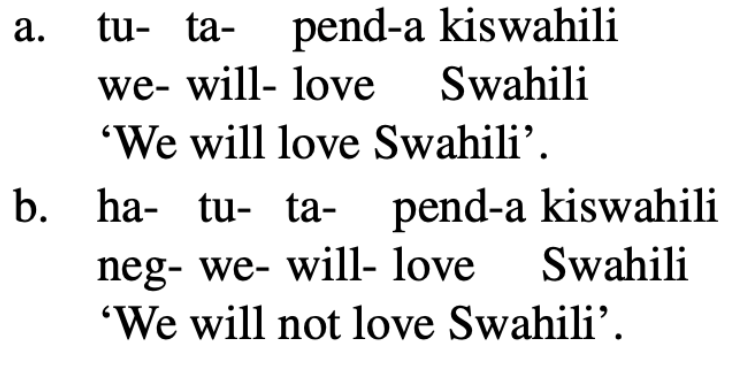}

    However, the same is not the case for 1SG subjects. In particular, 1SG ``I will love Swahili'' is expressed with the 1SG prefix in place of the 1PL in (a) above, negation cannot be expressed by an identical replacement of the 1SG into 1PL in (b) above. That expression of ``I will not love Swahili'' is ungrammatical---instead, the negation and subject 1SG must be expressed together in the single morpheme {\em si-}. These three cases in 1SG are demonstrated as (a-c) below:

     \includegraphics[width = .39\textwidth]{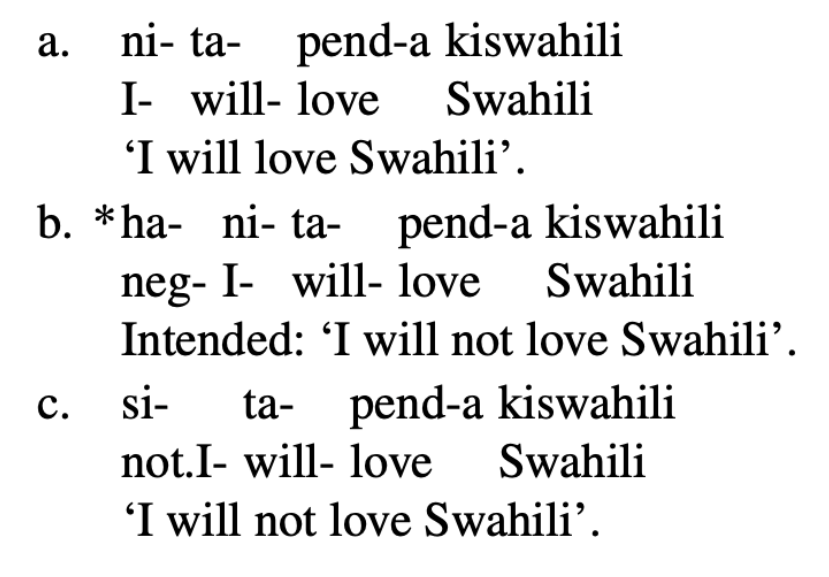}
}\end{ex}

The process of fusion is depicted in \cite{bobaljik_if_1997} in the following way, 
where the structure on the left has the two leaves $\mathtt{AGR}$ and $\mathtt{T}$ 
merged into one:  
\begin{center}
    \begin{tabular}{c c c c c}
   \begin{forest}
    [ $\mathtt{AGR}$ [ $\mathtt{V}$ ][$\mathtt{AGR}$ [$\mathtt{AGR}$\\{[$ \alpha, \beta, \gamma$]}] [ $\mathtt{T}$ \\{[$ \delta, \epsilon$]}]]]
\end{forest}  & $\longrightarrow$ & \begin{forest}
[$\mathtt{AGR}$ [ $\mathtt{V}$ ] [ {$\mathtt{AGR}/\mathtt{T}$} \\{[$ \alpha, \beta, \gamma, \delta, \epsilon$]}]] \end{forest}  
\\
\end{tabular}
\end{center}
The label $\mathtt{AGR/T}$ represents the fact that, with this formulation, it is unclear 
how the new leaf that has fused the two original leaves with different heads should be labeled. We will see below that this labeling issue can in fact be easily resolved.

We show that fusion can be formulated in our setting in a way that does \textit{not} require
modifying the underlying tree structure. 
We depict Fusion then as the following, where the feature bundles are being represented as their hierarchical tree structures.

\begin{center}
    \begin{tabular}{c c c}
   \begin{forest}for tree=nice empty nodes
    [$\mathtt{AGR}$ [$\mathtt{V}$][$\mathtt{AGR}$ [$\mathtt{AGR}$ \\{[$ \alpha, \beta, \gamma$]} [$\alpha$] [{[$\beta, \gamma$]}[$\beta$] [$\gamma$]]] [$\mathtt{T}$ \\{[$\delta,\epsilon$]}[$\delta$] [$\epsilon$]]]]
\end{forest}  & $\longrightarrow$ & \begin{forest}for tree=nice empty nodes
[ $\mathtt{AGR}$ [$\mathtt{V}$] [$\mathtt{AGR}$ \\{[$ \alpha, \beta, \gamma, \delta, \epsilon$]} [$\mathtt{AGR}$ \\{[$ \alpha, \beta, \gamma$]} [$\alpha$] [{[$\beta, \gamma$]}[$\beta$] [$\gamma$]]] [$\mathtt{T}$\\{[$\delta,\epsilon$]}[$\delta$] [$\epsilon$]]]]  \end{forest}  
\end{tabular}
\end{center}

The idea here is that we reassemble the morphosyntactic tree
by inserting an enlarged morphological tree at a leaf of a reduced syntactic tree, provided a matching condition holds, according to the matching rule $\Gamma_{SM}$ of Definition~\ref{featurematch}.

This means that, more generally, we represent fusion as the operation that transforms
a morphosyntactic tree of the form
$$ \Tree[ $T$ [ .$v$ \qroof{morphological tree $T_1$}.$(B_{v_1},\alpha_1)$   \qroof{morphological tree $T_2$}.$(B_{v_2},\alpha_2)$ ] ] $$
into a morphosyntactic tree of the form
$$ \Tree[ $T$ [ .$(B_v=B_{v_1}\cup B_{v_2},\alpha_v)$  \qroof{morphological tree $T_1$}.$B_{v_1}$  \qroof{morphological tree $T_2$}.$B_{v_2}$ ] ]  $$
where the vertex $v$, that is part of a syntactic object in the first tree, becomes part of a morphological object in
the second tree, with the associated feature bundle $B_v=B_1\cup B_2$. Note that the vertices $v_1$ and $v_2$,
in the first tree, are the vertices where the insertion operation $\gamma_{\cS\cO,\cM\cO}$ has attached
morphological data to the leaves of a syntactic tree, hence they are also decorated with the corresponding
data $\alpha_1,\alpha_2\in \cS\cO_0$ that were at the leaves of the syntactic tree, with the condition that
$(B_{v_i},\alpha_i)\in \Gamma_{SM}$ so that the insertion $\gamma_{\cS\cO,\cM\cO}$ can take place. 
In the second tree, the vertices $v_1,v_2$ become interior vertices of a morphological tree, hence they still carry
the $B_{v_i}$ labels but they no longer carry the syntactic $\alpha_i$ labels. On the other hand, the vertex $v$
has now become a leaf for the syntactic tree and the place where the insertion $\gamma_{\cS\cO,\cM\cO}$
takes place, so it needs to carry also a datum $\alpha_v\in \cS\cO_0$. The way to obtain this label is via the
head function $h_{T'}$ of the syntactic tree $T'$ and the labeling algorithm for syntactic objects. Note that
our choice to label the resulting syntactic vertex with $\alpha_v=\alpha_{h_{T'}(v)}$ seems to completely remove
the information coming from the other $\alpha_i$ that does not project. This is not the case, however, since we
must have the condition $(B_{v_1}\cup B_{v_2},\alpha_v)\in \Gamma_{SM}$ in order to still perform the matching
at $v$ of syntactic and morphological data. In the tree we start with, we have $(B_{h_{T'}(v)},\alpha_{h_{T'}(v)})\in 
\Gamma_{SM}$ and also $(B_i,\alpha_i)\in \Gamma_{SM}$ for the other leaf, but these two separate
conditions do not a priori imply $(B_{v_1}\cup B_{v_2},\alpha_v)\in \Gamma_{SM}$. The fact that this holds
(which is a necessary condition for fusion to take place) can be interpreted as a relation between 
$\alpha_1,\alpha_2$ that makes it possible to match both $B_{v_1}$ and $B_{v_2}$ to $\alpha_v$.

\begin{defn}\label{fusion}
The fusion operation is a linear operator $\cF: \cV(\cW_{\cM\cS}) \to \cV(\cW_{\cM\cS})$
on the space of morphosyntactic trees, extended by linearity and defined on basis elements
as $\cF(F)=\sqcup_i \cF(T_i)$ for $F=\sqcup_i T_i\in \cW_{\cM\cS}$, where
for $T\in \cM\cS$ we define $\cF(T)$ in the following way.
The morphosyntactic tree is of the form $$ T=\gamma_{\cS\cO,\cM\cO}(T',(S_\ell)_{\ell\in L(T')}), \ \ \text{ for }
T'\in \cS\cO \ \ \text{ and } S_\ell \in \widetilde{\cM\cO}. $$
We define
\begin{equation}\label{cherries}
{\bf C}(T): =\{ T_v \in {\rm Acc}(T) \,|\, T_v=\Tree[.$v$  \qroof{tree $S_1$}.{$(B_{v_1}, \alpha_1)$} \qroof{tree $S_2$}.{$(B_{v_2},\alpha_2)$} ]  \}\, 
\end{equation}
namely the set of accessible terms of the morphosyntactic trees that are cherry trees $\fM(\alpha_1,\alpha_2)$ 
of a syntactic tree with morphological insertions at both leaves, with $(B_{v_i},\alpha_i)\in \Gamma_{SM}$. 
We then set
\begin{equation}\label{fusionF}
\cF(T) =\sum_{T_v \in {\bf C}(T)} \gamma_{\cS\cO,\cM\cO}(T/^c \fM(\alpha_1,\alpha_2), S_{12}, (S_\ell)_{\ell\in L(T')\smallsetminus \{ \ell_1, \ell_2 \}} )\, ,
\end{equation}
where $\ell_1, \ell_2$ are the leaves of $T'$ marked by $\alpha_1,\alpha_2 \in \cS\cO_0$ and $S_{12}\in \widetilde{\cM\cO}$
is given by
\begin{equation}\label{S12}
S_{12} = \ \ \ \  
\Tree[ .$(B_v=B_{v_1}\cup B_{v_2},\alpha_v)$  \qroof{tree $S_1$}.$B_{v_1}$  \qroof{tree $S_2$}.$B_{v_2}$ ]   
\end{equation}
where $\alpha_v =\alpha_{h_{T'}(v)}\in \{ \alpha_1, \alpha_2 \}$. 
\end{defn}

Note that, here and in the following, we write for
simplicity $S_1$ and $S_2$ as the morphological objects
fusing to $S_{12}$, though they can just be any pair 
$S_\ell$, $S_{\ell'}$ attached to the same
syntactic vertex $v$ in the morphosyntactic tree.

\smallskip

The expression $\cF(T)$ of \eqref{fusionF} has the effect of removing the accessible term $\fM(\alpha_1,\alpha_2)$
of the syntactic object $T'$, leaving a leaf in place of the root vertex $v$ of $\fM(\alpha_1,\alpha_2)$ (which means
using the quotient $T'/^c \fM(\alpha_1,\alpha_2)$ that shrinks $\fM(\alpha_1,\alpha_2)$ to its root vertex, and then,
instead of $T=\gamma_{\cS\cO,\cM\cO}(T', S_1, S_2, \ldots, S_n)$, one computes
$\gamma_{\cS\cO,\cM\cO}(T'/^c \fM(\alpha_1,\alpha_2), S_{12}, S_3, \ldots, S_n)$. This performs the change that
we described above, and the sum over ${\bf C}(T)$ means that one considers (as a sum) all the possibilities where
this operation can be performed in the given $T$ (and similarly for a whole workspace $F\in \cW_{\cM\cS}$). 

Just as for Merge in syntax we can decompose $\cK=\sum_{S,S'} \fM_{S,S'}$ and for the
operations that assemble morphosyntactic trees we can decompose $\cK_T=\sum_{S_1,\ldots, S_n} \fM^T_{S_1,\ldots,S_n}$,
as discussed above, we can also similarly decompose
\begin{equation}\label{Fdecomp}
\cF = \sum_{S_1,S_2} \cF_{\gamma_{\cS\cO,\cM\cO}(\fM(\alpha_1,\alpha_2), S_1,S_2)}\, , 
\end{equation}
where the transformations $\cF_{\gamma_{\cS\cO,\cM\cO}(\fM(\alpha_1,\alpha_2), S_1,S_2)}$ 
target a specific morphosyntactic subtree given by the insertion
$\gamma_{\cS\cO,\cM\cO}(\fM(\alpha_1,\alpha_2), S_1,S_2)$ of two morphological trees $S_1,S_2$
with root feature bundles $B_{v_1}$ and $B_{v_2}$ into the two leaves of a 
syntactic object $\fM(\alpha_1,\alpha_2)$ with $(B_{v_i},\alpha_i)\in \Gamma_{SM}$. This map acts on
a workspace $F \in \cW_{\cM\cS}$ by searching for a copy of the object 
$\gamma_{\cS\cO,\cM\cO}(\fM(\alpha_1,\alpha_2), S_1,S_2)$ among the accessible terms of the
workspace, and replacing it with the object $\gamma_{\cS\cO,\cM\cO}((B_v,\alpha_v), S_{12})$
where $(B_v,\alpha_v)$ is the root vertex $v$ of $\fM(\alpha_1,\alpha_2)$ with this labeling. 
This can also be reformulated through a characterization of the fusion transformation in the following way.

\begin{prop}\label{Fusediagr}
The fusion operation $\cF$ of \eqref{fusionF} and \eqref{Fdecomp} can be characterized as the unique
map that makes the following diagram commute:
$$ \xymatrix{ \cV(\cW_\cM)  \ar[d]_{\fM^{\rm morph}_{S_1,S_2}} \ar[rr]^{\fM^T_{S_1,S_2,\ldots S_n}} & & \cV(\cW_{\cM\cS})
\ar@{-->}[d]^{\cF_{\gamma_{\cS\cO,\cM\cO}(\fM(\alpha_1,\alpha_2), S_1,S_2)}}  \\
\cV(\cW_\cM)   \ar[rr]^{\fM^{T/^c \fM(\alpha_1,\alpha_2)}_{S_{12},\ldots S_n}}  & &  \cV(\cW_{\cM\cS})
} $$
\end{prop} 

We can interpret Proposition~\ref{Fusediagr} as providing a
different but equivalent viewpoint on the fusion operation
$\cF$. The description we gave in terms of the map 
$\cF_{\gamma_{\cS\cO,\cM\cO}(\fM(\alpha_1,\alpha_2), S_1,S_2)}: \cV(\cW_{\cM\cS})\to \cV(\cW_{\cM\cS})$ is a transformation
of morphosyntactic trees. The corresponding map $\fM^{\rm morph}_{S_1,S_2}$ describes what happens if one sees fusion
as a transformation of morphological trees (as it is often described in DM). The two are related via the morphosyntax-assembly operations $\fM^T_{S_1,S_2,\ldots S_n}$ and $\fM^{T/^c \fM(\alpha_1,\alpha_2)}_{S_{12},\ldots S_n}$. This is one way
of reading the commutative 
diagram of Proposition~\ref{Fusediagr} (in other
words, to read it ``horizontally": the horizontal maps relate 
``fusion as an operation in morphology" and ``fusion as an operation in morphosyntax"). However, there is another way of reading the same diagram, namely reading it ``vertically".
When seen in this way, the two vertical arrows are a transformation between the two horizontal arrows, or in other words a
map that changes a morphosyntax-assembly operation
$\fM^T_{S_1,S_2,\ldots S_n}$ into another one,
$\fM^{T/^c \fM(\alpha_1,\alpha_2)}_{S_{12},\ldots S_n}$. This description shows that we can also think of fusion as
acting on the set of operations $\{ \fM^T_{S_1,\ldots, S_n} \}$.
As we will see below, this is the case also for the other
operations of DM.

\subsection{Fission} \label{FissionSec}

Fission is the process of splitting one feature bundle into two. We can motivate the existence of fission by the fact that there is reason to believe that two features exist in one single feature bundle within the syntax, but are realized as separate phonological exponents. Consider the following data from Arabic, as given by \cite{hewett_allomorphy_2023} originally from \cite[p.56]{watson1993syntax}.

\begin{ex}{\rm
    \d{S}an\textrevglotstop\=an\=\i  \text{ } Arabic has discontinuous agreement of person and number in the context of prefix conjugation. With the standard assumption in syntax that the subject's morphological features are all housed in a single syntactic head, which includes both person and number, only first person consistently has these features realized as a single affix (the prefix). Second and third person exemplify discontinuous agreement: the person and number are realized as both prefixes and suffixes on the verb. For example, the verb {\em gmbr}, `sit', is conjugated as follows, where the person agreement affixes are indicated in bold:

\begin{center}
\begin{tabular}{l l l}
& SG & PL\\
\hline
    1 & \textbf{\textglotstop a}-gambir & \textbf{ni}-gambir \\
    2M & \textbf{ti}-gambir & \textbf{ti}-gambir-\textbf{\=u} \\
    2F & \textbf{ti}-gambir-\textbf{\=\i} & \textbf{ti}-gambir-\textbf{ayn} \\
    3M & \textbf{yi}-gambir & \textbf{yi}-gambir-\textbf{\=u} \\
    2F & \textbf{ti}-gambir & \textbf{yi}-gambir-\textbf{ayn}
\end{tabular}
\end{center}
}\end{ex}

\smallskip 

The formulation of \cite{hewett_allomorphy_2023} 
presents the idea that the two subsets of features to be fissioned from each other are each partitioned 
into a separate feature bundle, but the remaining features, indicated by $\phi$, are copied into \textit{both} 
fissioned bundles. There is reason to think that the 
non-fissioned features exist in both places, because they sometimes are pronounced multiple times, in each of the two vocabulary items corresponding to the two feature bundles resulting from fission. Hence $\phi$, the remaining features irrelevant for fission, should appear in both feature bundles.

Suppose that $\alpha$ and $\beta$ are to be fissioned from $\gamma$. This can be depicted as follows, where $\phi$ represents one or more additional features: 

\begin{center}
    \begin{tabular}{c c c}
   \begin{forest}
    [ $\mathtt{T}$ [$\mathtt{ASP}$ ] [ $\mathtt{T}$ \\{[$ \alpha, \beta, \gamma, \phi$]}] ]
\end{forest}  & $\longrightarrow$ & \begin{forest}
[ $\mathtt{T}$ [ $\mathtt{ASP}$ ] [ $\mathtt{T}$ [ $\mathtt{T}$\\{[$ \alpha, \beta,\phi$]}] [ $\mathtt{T}$ \\{[$\gamma,\phi$]}]]] \end{forest}  \\
\end{tabular}
\end{center}

If we depict this with our tree interpretation of feature bundles, this becomes the following example.

\begin{ex}\label{FissionEx}{\rm 
A fission operation (with fissioned feature $\phi$) is illustrated by the example:
\begin{center}
    \begin{tabular}{c c c}
   \begin{forest}
    [ $\mathtt{T}$ [ $\mathtt{ASP}$ ] [ $\mathtt{T}$ \\{[$ \alpha, \beta, \gamma,\phi$]} [$\alpha$] [{[$ \beta, \gamma, \phi$]} [$\beta$] [{[$\gamma, \phi$]}[$\gamma$] [$\phi$]]] ]]
\end{forest}  & $\longrightarrow$ & \begin{forest}
[ $\mathtt{T}$ [ $\mathtt{ASP}$ ] [ $\mathtt{T}$ [ $\mathtt{T}$ \\{[$ \alpha, \beta, \phi$]} [$\alpha$] [{[$\beta, \phi$]}[$\beta$] [$\phi$]]] [$\mathtt{T}$ \\{[$\gamma,\phi$]} [$\gamma$][$\phi$]]]] \end{forest}  \\
\end{tabular}
\end{center}}
\end{ex}

As observed in Remark~\ref{unionBv}, we can realize a bundle $B_v$ of features at the root of a
morphological tree $T\in \cM\cO$ through alternative tree structures where some of the features $\phi\in B_v$
are repeated in lower vertices of the tree not on the same root-to-leaf path. The fission operations 
can be seen as transformations that alter the tree decomposition of a certain bundle of morphological
features, while at the same time replacing a morphological vertex with a syntactic one, hence lowering
the boundary between syntax and morphology, unlike fusion which raises it. We will discuss this more in \S \ref{BoundarySec}.

Again we define the fission operations as linear transformations on the space $\cV(\cW_{\cM\cS})$ of
morphosyntactic trees, by defining the action on a single morphosyntactic tree, extending it multiplicatively
(in the $\sqcup$ product) on forests and additively on linear combinations, as we did for the case of fusion.
Also, as in the case of fusion, we define the operator that performs all the possible fission operations on
a given tree (or forest) and presents the results as a sum of each individual fission (which is the analog
of the $\cF(T)$ fusion of \eqref{fusionF}), and then we decompose it into individual fission operations
(as in \eqref{Fdecomp} for the case of fusion). 

Consider a morphological tree $S\in \widetilde{\cM\cO}$ and let $B$ be the 
bundle of morphological features at the root vertex $v$ of $S$. For any subset of
this feature bundle, $A\in \cP(B)$, consider the set $B\smallsetminus A$ (the subset of $B$ not including any elements of the set $A$) and the
set of all partitions 
\begin{equation}\label{PAB}
 \fP_{A,\alpha_1,\alpha_2}(B):= \{ (B_1, B_2)\,|\, B\smallsetminus A = B_1 \sqcup B_2 \text{ and } (B_i\cup A,\alpha_i)\in \Gamma_{SM} \}  \, . 
\end{equation}

\begin{defn}\label{fissiondef}
Consider a given $T\in \cM\cS$ with $T=\gamma_{\cS\cO,\cM\cO}(T',S_1,\ldots, S_n)$ for $T'\in \cS\cO$ and
$S_\ell \in \widetilde{\cM\cO}$, for $\ell\in L(T')$, with $(B_{v_\ell},\alpha_\ell)\in \Gamma_{SM}$ for
$v_\ell$ the root vertex of $S_\ell$ that $\gamma_{\cS\cO,\cM\cO}$ inserts at the $\ell$ leaf of $T'$. 
We define the fission operation as
\begin{equation}\label{fissionPhi}
\Phi(T)=\sum_{\ell,\, \alpha, \, A, \, (B_1,B_2)} \Phi_{A,(B_1,B_2),\alpha}(T) \ \ \ \text{ with } \begin{array}{ll} \ell\in L(T'), \ \ \alpha\in \cS\cO_0 , \\ A\in \cP(B_{v_\ell}), \ (B_1,B_2)\in \fP_{A,\alpha_\ell, \alpha} (B_{v_\ell}), \end{array}
\end{equation}
\begin{equation}\label{fissionPhi2}
\Phi_{A,(B_1,B_2),\alpha}(T)= \gamma_{\cS\cO,\cM\cO}(
T'  \circ_{\cS\cO_\ell} \fM(\alpha_\ell, \alpha),  S_1,\ldots, S_{\ell, B_1\cup A}, S_{\ell, B_2\cup A},
, \ldots, S_n)\, , 
\end{equation}
where the operation $\circ_{\cS\cO_\ell}$ is the single leaf insertion as in \eqref{circellSO}, and
the two morphological trees $S_{\ell, B_i\cup A}$ are obtained from $S_\ell$
by performing the following operations:
\begin{itemize}
\item replacing $B_{v_\ell}$ at the root with $B_{v_i}=B_i \cup A$,
\item for each vertex $w$ of $S_\ell$ below the root, take $B_w\cap (B_i \cup A)$
\item in the resulting tree consider the subforest $F_{\underline{w}}$ whose components $S_{w_i}$
are the accessible terms of $S_\ell$ with root vertex $w_i$ (and hence all other vertices as well) satisfying 
$B_{w_i}\cap (B_i \cup A)=\emptyset$ 
\item take $S_{\ell, B_i\cup A} =S_\ell /^\rho F_{\underline{w}}$ with the vertices $w$ labelled by
$B_w\cap (B_i \cup A)$.
\end{itemize}
Since $\fM(\alpha_\ell,\alpha)$ is non-planar, in the insertion of $S_{\ell, B_1\cup A}, S_{\ell, B_2\cup A}$ at the leaves
of $\fM(\alpha_\ell, \alpha)$ both possibilities (differing in the assignment of the head function) are included in the sum:
$$ \Tree[ \qroof{S_{\ell, B_1\cup A}}.{$(B_1\cup A, \alpha_\ell)$} \qroof{S_{\ell, B_2\cup A}}.{$(B_2\cup A, \alpha)$} ] \ \ \ \text{ and } \ \ \ 
\Tree[ \qroof{S_{\ell, B_1\cup A}}.{$(B_1\cup A, \alpha)$} \qroof{S_{\ell, B_2\cup A}}.{$(B_2\cup A, \alpha_\ell)$} ]  $$
\end{defn}

The procedure described here is the general form of what we have seen in Example~\ref{FissionEx}. 

\begin{ex}\label{fissex2}{\rm
For example, consider the case of a morphological tree of the form
$$   S=\Tree[ .{$[\phi, \alpha, \beta, \gamma]$} [ .{$[\phi, \alpha]$} $\phi$ $\alpha$ ] [ .{$[\beta, \gamma]$}
  $\beta$ $\gamma$  ] ]  $$
  and take $B_1\cup A=[\phi, \gamma]$ and $B_2\cup A=[\phi, \alpha, \beta]$ with $A=\phi$. The procedure described above
  for the construction of the trees $S_{B_i\cup A}$ starts with producing the list of feature bundles $B_w\cap (B_i\cup A)$,
$$ \begin{array}{cccc} 
& B_w\cap (B_1\cup A) &  B_w\cap (B_2\cup A) \\
\text{level 2} &  \left[ \phi , \alpha \right] \cap [\phi, \gamma]= \phi &   [\phi, \alpha]\cap [\phi, \alpha, \beta] =[\phi, \alpha] \\
\text{level 2} & \left[\beta, \gamma \right] \cap [\phi, \gamma]=\gamma &  [\beta, \gamma] \cap [\phi, \alpha, \beta] = \beta \\
\text{level 3} & \phi \cap [\phi, \gamma]=\phi \ \ \alpha \cap [\phi, \gamma ]= \emptyset & \phi \cap [ \phi, \alpha, \beta] = \phi \ \ \  \alpha \cap [ \phi, \alpha, \beta] =\alpha \\
\text{level 3} & \beta \cap [\phi, \gamma]=\emptyset \ \ \gamma \cap [\phi, \gamma]=\emptyset &   \beta \cap [ \phi, \alpha, \beta] = \beta \ \ 
\gamma \cap [ \phi, \alpha, \beta] =\emptyset
\end{array} $$
which gives the trees
$$ S_{B_1\cup A} = \Tree[ .{$[\phi, \gamma]$} [.$\phi$ $\phi$ ] $\gamma$ ] \ \ \ \text{ and } \ \ \  S_{B_2\cup A}=\Tree[ .{$[\phi, \alpha, \beta]$}
[ .{$[\phi, \alpha]$} $\phi$ $\alpha$ ] [.$\beta$ $\beta$ ] ] $$
The two non-branching vertices can be eliminated as they
do not add any features, resulting in the trees
$$ S_{B_1\cup A} = \Tree[ .{$[\phi, \gamma]$}  $\phi$ $\gamma$ ] \ \ \ \text{ and } \ \ \  S_{B_2\cup A}=\Tree[ .{$[\phi, \alpha, \beta]$}
[ .{$[\phi, \alpha]$} $\phi$ $\alpha$ ] $\beta$ ] $$}
\end{ex}

In order to characterize fission with a commutative diagram akin to the commutative diagram we gave for fusion, we introduce the \textit{root-cut operator}.

\begin{defn}\label{rootcut}
The root-cut operator from trees to forests is defined by
\begin{equation}\label{fCT}
\fC(T)=T_1 \sqcup \cdots \sqcup T_n\, , \ \ \ \text{ for } T=\fB(T_1 \sqcup \cdots \sqcup T_n) \, .
\end{equation}
Namely, it performs the opposite operation of the grafting $\fB$: instead of appending all the
component trees of a forest to a common root, forming a single tree, it cuts all the edges 
below the root of a tree, resulting in a forest. In particular, if 
$\fT_X$ is the set of non-planar binary rooted trees with leaves labelled by elements of a set $X$ and $\fF_X$ is the
set of forests with components in $\fT_X$, 
\begin{equation}\label{fCT2}
\fC(T)=T_1 \sqcup T_2\, , \ \ \ \text{ for } T=\fM(T_1,T_2)=\fB(T_1\sqcup T_2) \, .
\end{equation}
\end{defn}

In particular, for the case of morphological trees, we write 
$$ \fC_S^{\rm morph}: \cV(\cW_\cM) \to \cV(\cW_\cM) $$
for the map of morphological workspaces that acts as the
root-cut operation on a component $S$ of the workspace
and as the identity on the other components,
\begin{equation}\label{CutMorph}
  \fC_S^{\rm morph}(S \sqcup F)=\fC(S)\sqcup F = S_1 \sqcup S_2 \sqcup F, \ \ \  \text{ for } S=\fM^{\rm morph}(S_1,S_2) \, . 
 \end{equation} 
 We consider the following family of maps of a similar type:
 \begin{equation}\label{CutMorphSi}
 \fC^{S}_{A, (B_1,B_2)}(S \sqcup F)= S_{B_1\cup A} \sqcup S_{B_2\cup A} \sqcup F\, ,
 \end{equation}
 for $S_{B_1\cup A}, S_{B_2\cup A}$ constructed as above.

We can then give a characterization of fission similar to the one we gave of fusion in Proposition~\ref{Fusediagr}.

\begin{prop}\label{fissdiagr}
The fission operation $\Phi$ of \eqref{fissionPhi} and \eqref{fissionPhi2} can be characterized as the unique
map that makes the following diagram commute:
$$ \xymatrix{ \cV(\cW_\cM)  \ar[d]_{\fC^{S}_{A, (B_1,B_2)}} \ar[rrrr]^{\fM^T_{S_1,\ldots, S_\ell, \ldots S_n}} &  & &  & \cV(\cW_{\cM\cS})
\ar@{-->}[d]^{\Phi_{A,(B_1,B_2),\alpha}}  \\
\cV(\cW_\cM)   \ar[rrrr]^{\fM^{T\circ_\ell \fM(\alpha_\ell,\alpha)}_{S_1,\ldots, S_{\ell, B_1\cup A}, S_{\ell, B_2\cup A}, \ldots S_n}}  & & & &  \cV(\cW_{\cM\cS})
} $$
In the case where $A=\emptyset$ and $B_1, B_2$ are the labels of the vertices $v_1, v_2$ of the subtrees $S_{v_1}, S_{v_2}$
with $S_\ell=\fM^{\rm morph}(S_{v_1}, S_{v_2})$, the arrow $\fC^{S}_{A, (B_1,B_2)}$
is just the same as the root cut $\fC^{\rm morph}_S$ of \eqref{fCT}, in the form \eqref{CutMorph}. 
\end{prop}

Again Proposition~\ref{fissdiagr} provides us with two equivalent
interpretations of fission: one as discussed above, as transformations of morphosyntactic and morphological trees, and the other as transformations acting on the set of operations $\{ \fM^T_{S_1,\ldots,S_n} \}$ of morphosyntax formation.

\subsection{Obliteration} \label{OblImpSec} 

Obliteration is the complete removal of a feature bundle. 

\begin{ex}\label{ondarru} {\rm
As discussed in \cite{arregiMorphotacticsBasque2012},  the Ondarru dialect of Basque displays obliteration. Specifically, a first-person clitic is deleted when followed by a first- or second-person ergative clitic. This is described in \cite{keine_impoverishment_nodate} as the following rule:

    \begin{itemize}
        \item Within an auxiliary M-word with two clitics, $c_1$ and $c_2$, delete $c_1$ when $c_1$'s feature bundle has [+participant, +author] and $c_2$'s feature bundle has [ergative, +participant].
    \end{itemize}
   
}\end{ex}

\begin{rem}\label{Onda} {\rm
The high frequency of these DM operations as being triggered by particular (i.e., language-specific) combinations of features suggests that these may correspond to filtering at the Externalization phase by language-dependent coloring rules. For example, one Externalization coloring rule pertaining to Ondarru's rule given in \ref{ondarru} would be to filter out any tree structures where the two colors $\mathfrak{c}_1$ and $\mathfrak{c}_2$ corresponding to $c_1$ and $c_2$ are adjacent/part of the same auxiliary M-word and $c_1$ and $c_2$'s feature bundles contained [+participant, +author] features and [ergative, +participant] features, respectively.
} \end{rem}

Obliteration hence corresponds to removing an entire 
morphological feature tree. This is motivated by any morphological case where an entire feature bundle is realized phonologically in some cases, but is not realized at all in others. In this case one can posit that the feature bundle was deleted and hence inaccessible to the usual vocabulary insertion rules that would normally realize that feature bundle, in other (morphosyntactic) circumstances. 

In order to model obliteration within our formalism,
we need to consider operations of the form 
$\fM^T_{S_1,\ldots,S_n}$, where one or more
of the $\{ S_\ell \}_{\ell \in L(T)}$ are {\em empty}.
This corresponds to cases where a syntactic leaf plays
a role in the syntactic structure but does not carry
an associated bundle of morphological features.

\begin{rem}\label{GammaSMnosurj}{\rm
Allowing for some empty morphological insertions $S={\bf 1}$
requires weakening the assumptions that we made regarding
the matching rules, specified by the correspondence $\Gamma_{SM}$
of Definition~\ref{featurematch}, by dropping the surjectivity assumption that $\pi_2 |_{\Gamma_{SM}} : \Gamma_{SM} \twoheadrightarrow \cS\cO_0$. Indeed, if the map 
$\pi_2 |_{\Gamma_{SM}}$ is not surjective, the set
$\cS\cO_0\smallsetminus \pi_2(\Gamma_{SM})$ represents
the set of lexical items and syntactic features at which
obliteration can happen. 
}\end{rem}

Let us denote by ${\bf 1}$ the unit of the magma of
morphological objects, namely the formal empty tree. This
is also the unit of the multiplication $\sqcup$ of the
Hopf algebra of morphological workspaces. We can then
formalize obliteration in the following wsy.

\begin{prop}\label{oblitprop}
Obliteration $\bO_S: \cV(\cW_{\cM\cS})\to \cV(\cW_{\cM\cS})$
acts by 
\begin{equation}\label{bOS}
\bO_S(\gamma_{\cS\cO,\cM\cO}(T,S,S_1,\ldots,S_n)) := 
\gamma_{\cS\cO,\cM\cO}(T,{\bf 1},S_1,\ldots,S_n))= \fM^T_{{\bf 1}, S_1,\ldots, S_n}(F)\, ,
\end{equation}
where $F$ is the morphological 
workspace $F=S\sqcup S_1\sqcup \cdots \sqcup S_n$. 
\end{prop}

\proof
We can view the feature bundle to be obliterated 
as a component $S$ of a morphological workspace 
$F=S\sqcup \hat F=S\sqcup S_1\sqcup \cdots \sqcup S_n$, 
and we can assume that we start with a morphosyntactic
tree of the form $\gamma_{\cS\cO,\cM\cO}(T,S,S_1,\ldots,S_n)$, 
where the components of the workspace $F$ are inserted
at the leaves of a syntactic tree $T$ through the
action of the morphosyntactic building operation
$$ \sqcup \circ (\gamma_{\cS\cO,\cM\cO}(T,\ldots) \otimes{\rm id} \circ \delta_{\underline{B},\underline{\alpha}} \circ \Delta^\rho $$
restricted to the term where $\gamma_{\cS\cO,\cM\cO}(T,\ldots)$ acts on the primitive term $F\otimes 1$ of the coproduct. 

In order to accommodate the obliteration of $S$, 
notice that 
the primitive part of the coproduct performs all the partitions of the workspace. In particular, there will be a term
in the primitive part of the coproduct that is of the form
$\hat F \otimes S=S_1\sqcup \cdots \sqcup S_n \otimes S$.
We have
$S_1\sqcup \cdots \sqcup S_n={\bf 1}\sqcup S_1\sqcup 
\cdots \sqcup S_n$, hence we can formally apply the insertion
operation $\gamma_{\cS\cO,\cM\cO}(T,\ldots)$ to the term
$\hat F \otimes S$ of the coproduct. This means performing
the operation $\fM^T_{{\bf 1},S_1,\ldots, S_n}$. 
\endproof

Note that the operation $\fM^T_{{\bf 1},S_1,\ldots, S_n}$
is an analog here of the operation $\fM_{{\bf 1},S}$ used in
syntax as a piece of the operation needed to model 
Internal Merge (see \S 1.4.3 of \cite{marcolli_mathematical_2024}).

Again, we write for simplicity the substitution $S\mapsto {\bf 1}$
in the first position of $(S,S_1,\ldots, S_n)$, but in fact it can be at any position $S_\ell$ corresponding to any leaf $\ell\in L(T)$.

\begin{cor}\label{OblMT}
We can reinterpret obliteration, defined as in \eqref{bOS},
as the operation that maps $\bO_S: \fM^T_{S,S_1,\ldots, S_n} \mapsto \fM^T_{{\bf 1}, S_1,\ldots, S_n}$.
\end{cor}

\proof This follows directly from Proposition~\ref{oblitprop} by
writing in \eqref{bOS} 
$$ \gamma_{\cS\cO,\cM\cO}(T,S,S_1,\ldots,S_n) = \fM^T_{S,S_1,\ldots, S_n}(F) $$
for any workspace $F$ that has a term of the form $S\sqcup S_1\sqcup \cdots \sqcup S_n\otimes F'$, for some $F'$, in the
coproduct $\Delta^\rho(F)$. 
\endproof

\smallskip

With this formulation of obliteration, one should
further specify how to interpret the role of the
leaf of the syntactic tree where the obliterated
morphological $S$ was inserted, when $S \mapsto {\bf 1}$.
One possibility is that the leaf $\ell$ with its 
label $\alpha_\ell$ is also removed (with $T$
replaced by the maximal full binary tree $T/^d \ell$ 
remaining after the cut of $\ell$). The other
possibility is that no morphology is inserted but
the leaf $\alpha$ with its syntactic label 
$\alpha_\ell$ is maintained. The mathematical
formulation suggests that it should be the
second case, since in the expression 
$\gamma_{\cS\cO,\cM\cO}(T,{\bf 1}, S_1, \ldots, S_n)$ 
the term ${\bf 1}$ means that no morphology is inserted
leaving the leaf of $T$ unchanged, and no
operation $T/^d \ell$ is involved. This second case 
seems also preferable from the syntactic point of view, 
in terms of the No Complexity Loss principle for
syntactic objects as discussed in \S 1.6.3 of
\cite{marcolli_mathematical_2024}.

\smallskip

\subsubsection{Obliteration of a subbundle of features}\label{OblSubSec} 

More generally, obliteration can also apply to a part of 
a feature bundle as in the following example.

\begin{ex}\label{OblEx2} {\rm 
Suppose that we start with a morphosyntactic
tree $$ \gamma_{\cS\cO.\cM\cO}(T,S,S_1,\ldots, S_n)
= \fM^T_{S,S_1,\ldots, S_n} (F) $$
for a morphological workspace $F= S\sqcup S_1 \sqcup \cdots \sqcup S_n \sqcup F'$, and 
with a morphological tree $S$ of the form
\begin{equation}\label{exSobl}
   S=\Tree[ .{$[\phi, \alpha, \beta, \gamma,\delta]$} [ .{$[\phi, \alpha]$} $\phi$ $\alpha$ ] [ .{$[\beta, \gamma,\delta]$}
  $\beta$ [ .{$[ \gamma, \delta ]$} $\gamma$ $\delta$ ] ] ] 
\end{equation}
that is inserted at a leaf $v$ of a syntactic tree $T$, with $(B_v,\alpha_v)\in \Gamma_{SM}$ for
$B_v=[\phi, \alpha, \beta, \gamma,\delta]$. Perform first a fission operation that transforms the tree above into
$$ S_v=\Tree[ .{$v$} [ .{$[\phi, \alpha]$} $\phi$ $\alpha$ ] [ .{$[\beta, \gamma,\delta]$}
  $\beta$ [ .{$[ \gamma, \delta ]$} $\gamma$ $\delta$ ] ] ]  $$
   where now the insertion leaves are the vertices $v_1$ and $v_2$ below $v$ with
  $\alpha_1,\alpha_2$ with $(B_i,\alpha_i)\in \Gamma_{SM}$ for $B_1=[\phi, \alpha]$ and
  $B_2=[\beta, \gamma,\delta]$ and one of the $\alpha_i$ equal to $\alpha_v$, as discussed above. We can write this tree as
  $$ S_v = \gamma_{\cS\cO,\cM\cO}(\fM(\alpha_1,\alpha_2), S_{B_1}, S_{B_2}) \, , $$
with
\begin{equation}\label{SB1SB2ex}
S_{B_1} =\Tree[ .{$[\phi, \alpha ]$} $\phi$ $\alpha$  ] \ \ \ \text{ and } \ \ \ S_{B_2} =\Tree[ .{$[\beta, \gamma, \delta]$} $\beta$ [ .{$[\gamma, \delta]$} $\gamma$ $\delta$ ] ]  \, ,
\end{equation}
and the resulting morphosyntactic tree as
$$ \gamma_{\cS\cO,\cM\cO}(T \circ_v \fM(\alpha_1,\alpha_2),
S_{B_1}, S_{B_2}, S_1,\ldots, S_n) \, . $$
 Suppose then that the bundle of features $B_1$
  is the part that we want to obliterate and $B_2$
  the part we want to keep.
  In the coproduct $\Delta^\rho(S_v)$, there is a term corresponding
to the admissible cut that removes the edge 
connecting $v$ to $v_2$, which is of the form 
$$  \Tree[ .{$[\beta, \gamma, \delta]$} $\beta$ [ .{$[\gamma, \delta]$} $\gamma$ $\delta$ ] ]   
\ \ \otimes 
\Tree[ .$v$ [ .{$[\phi, \alpha ]$} $\phi$ $\alpha$  ] ]
 \, .  $$
There is then a term in the 
coproduct $\Delta^\rho(F)$ of the morphological workspace 
that is of the form
$$  \Tree[ .{$[\beta, \gamma, \delta]$} $\beta$ [ .{$[\gamma, \delta]$} $\gamma$ $\delta$ ] ]  \ \sqcup S_1 \sqcup \cdots \sqcup S_n \ \ \ \otimes \Tree[ .$v$ [ .{$[\phi, \alpha ]$} $\phi$ $\alpha$  ] ]  \sqcup F' $$
Applying
$$ \sqcup \circ (\gamma_{\cS\cO,\cM\cO}(T', \ldots) \otimes {\rm id}) \circ \Delta^\rho $$
to this term of the coproduct then gives the morphosyntactic
tree with the obliterated $S_{B_1}$.
} \end{ex}  

\begin{rem}\label{exImp1}{\rm 
For the same morphological tree of \eqref{exSobl},
where, as in Example~\ref{OblEx2}, we want to  
remove the features $\phi$ and $\alpha$ while keeping $\beta, \gamma, \text{ and } \delta$,
performing the fission operation is optional. Indeed,
we might as well directly apply the coproduct to $S$
and select the term of the form 
$$ \Tree[ .{$[\beta, \gamma,\delta]$}
  $\beta$ [ .{$[ \gamma, \delta ]$} $\gamma$ $\delta$ ] ] \otimes \ \   \Tree[ .{$[\phi, \alpha, \beta, \gamma,\delta]$} [ .{$[\phi, \alpha]$} $\phi$ $\alpha$ ] ] $$
When other components of the morphological workspace $S\sqcup_i S_i$ are also taken into consideration,
one has a corresponding term in $\Delta^\rho(F)$ of the form
$$ S' \sqcup_i S_i \otimes S/^\rho S' \sqcup F' $$
for $F=S \sqcup_i S_i \sqcup F'$, and with
$$ S':= \Tree[ .{$[\beta, \gamma,\delta]$}
  $\beta$ [ .{$[ \gamma, \delta ]$} $\gamma$ $\delta$ ] ] \ \ \ \text{ and } \ \ \  S/^\rho S' = \Tree[ .{$[\phi, \alpha, \beta, \gamma,\delta]$} [ .{$[\phi, \alpha]$} $\phi$ $\alpha$ ] ] $$
We can then form the morphosyntactic object 
$$ 
\gamma_{\cS\cO,\cM\cO}(T',S',S_1,\ldots,S_n) 
$$ 
by applying
$$ \sqcup \circ (\gamma_{\cS\cO,\cM\cO}(T',\ldots) \otimes {\rm id}) \circ \Delta^\rho $$
to this term of the coproduct.
} \end{rem}

The reason for introducing the fission operation in
Example~\ref{OblEx2} is that it also allows us to  
obliterate {\em any other combinations of features}, as
the following example shows.

 \begin{ex}\label{exImp2}{\rm  
 Suppose then that, in the same example of  \eqref{exSobl} we want instead to remove the features $\phi$ and $\gamma$
 and retain $[\alpha, \beta, \delta]$. This can now be done in a similar way, but it first requires using a fission operation
 that performs the separation of the sets of features $B_1=[\phi, \gamma]$ and $B_2=[\alpha, \beta, \delta]$, namely the
 operation $\Phi_{\emptyset,(B_1,B_2),\alpha_v}$ as in \eqref{fissionPhi2}, where the procedure of
 Definition~\ref{fissiondef} for the construction of the $S_{B_1\cup A}$ and $S_{B_2\cup A}$ gives a
 resulting tree
 $$ S_v =  \Tree[ .$v$ $S_{B_1}$ $S_{B_2}$ ] =
 \Tree[ .$v$ [ .{$[\phi, \gamma]$} $\phi$ $\gamma$ ] [ .{$[\alpha, \beta, \delta]$} $\alpha$  [ .{$[\beta, \delta]$} $\beta$ $\delta$ ] ] ] 
 $$
 The coproduct $\Delta^\rho$ will then produce a term $S_{v_2} \sqcup_i S_i \otimes S_v/^\rho S_{v_2}$ where
 $$ S_{v_2} =\Tree[ .{$[\alpha, \beta, \delta]$} $\alpha$  [ .{$[\beta, \delta]$} $\beta$ $\delta$ ] ]\, , \ \ \ \text{ and } \ \ \ 
  S_v/^\rho S_{v_2} =\Tree[ .$v$ [ .{$[\phi, \gamma]$} $\phi$ $\gamma$ ] ] \, . $$
 We then apply the insertion $\gamma_{\cS\cO,\cM\cO}(T, \ldots)$ to this term of the coproduct, as in
 the previous example.
 }\end{ex}

Cases like Example~\ref{OblEx2} and Example~\ref{exImp2}
are suitable for modeling situations where the presence
of certain other features cause some of the features to
be obliterated. This would mean that, in Externalization,
a (language-dependent) filtering selects between the 
morphosyntactic structure before or after the obliteration 
operation, depending on the combination of features
present in the feature bundle before obliteration.
This type of filtering suggests
a formulation in terms of coloring algorithms where
certain adjacent combinations of colors are ruled out,
as mentioned in Remark~\ref{Onda}.
We will not elaborate on this further in the present paper,
as filtering in Externalization needs to be modeled 
separately.

\begin{rem}\label{ImpvsObl}{\rm
The case discussed above, where obliteration targets
a subbundle of features, as in Example~\ref{OblEx2}
or Example~\ref{exImp2}, may be thought of either
as Obliteration (though applied to only a part of
the morphological tree) or as a
case of Impoverishment. The difference 
between Impoverishment and Obliteration is sometimes
described by the fact that in the first case there is still a realization of an existent feature bundle, which just does not include the normal features being realized, whereas Obliteration completely disallows any realization of a morphological node (and hence that node/feature bundle must have 
been completely deleted). We can also make 
the distinction in terms of whether a bundle or subbundle of features is completely obliterated, or whether a trace of it is maintained, either referring to the first as a form of Obliteration and the second as Impoverishment, or to both as two different forms of Impoverishment. We discuss these different cases in \S \ref{OblImpSec2}. We will give in Proposition~\ref{ImpovProp} a comparative formulation, viewing these possibilities as different cases of Impoverishment.
}\end{rem}

\smallskip
\subsection{Impoverishment} \label{OblImpSec2}

Impoverishment involves the obliteration of a piece of the feature bundle, as in the cases discussed in Examples~\ref{OblEx2} and
\ref{exImp2}, and in more general cases discussed below.

This occurs when there is reason to think multiple features (e.g.,~person and number) are normally realized, but in specific circumstances only some of those features are realized (e.g. only person, not number). An example of this can be seen in classical Arabic \cite{haywood1965arabic}, as presented by \cite{embick_distributed_2007}.

\begin{ex}{\rm
    In the substantival declension of classical Arabic, certain substantives in the genitive indefinite do not express both superior and indefinite features, but rather express the default suffix. The following is a table of Arabic declensions.\footnote{N, G and A indicate nominative, genitive and accusative cases, respectively, while I and D indicate indefinite and definite.}

\begin{center}
\begin{tabular}{l l l l l l l}
& NI & GI & AI & ND & GD & AD\\
\hline
 rajul- `man' & -u-n & -i-n & -a-n & -u & -i & -a \\
 rij\=al- `men' & -u-n & -i-n & -a-n & -u & -i & -a \\
 h\=ašim- `Hashim' & -u-n & -i-n & -a-n & && \\
  h\=ar\=un- `Aaron' & -u & -a & -a &  &  & \\
  mad\=a\textglotstop in- `cities' & -u & -a & -a & -u & -i & -a 
\end{tabular}
\end{center}

The realization of {\em h\=ar\=un} `Aaron' and {\em mad\=a\textglotstop in} `cities' in the genitive indefinite as compared to the other substantives as well as the genitive definite case is quite different--instead of being realized as {\em -i-n}, it is realized as {\em -a}. Positing /i/ as the vocabulary item for [+oblique], and /a/ as the elsewhere vocabulary item, as well as /n/ as the vocabulary item for [-definite] and $\emptyset$ as the elsewhere for definiteness, it appears that these two features ([+oblique] and [-definite]) have been impoverished and then realized as their elsewhere forms (/-a/ and $\emptyset$).
}\end{ex}

Since we have conceptualized obliteration in Proposition~\ref{oblitprop} in terms of the coproduct $\Delta^\rho$, as well as the cases discussed in Examples~\ref{OblEx2} and
\ref{exImp2}, the general form of impoverishment should also
 be a combination of fission and coproducts, to separate out the subtrees of features to be removed or kept, and to actually 
 remove (as in obliteration) the unwanted part.

In our model, this 
 would again correspond to fissioning the feature bundle into the piece to be obliterated and the piece to remain, and then obliterating that part of the feature bundle. If we 
 view impoverishment as modeling cases where
spell-out at a terminal node, by vocabulary insertion
determined by specific features, is blocked by other less specific vocabulary items, then the difference
with obliteration can be seen as maintaining the
presence of certain features (and their possible interaction with other features, for instance in terms of determining
filtering in Externalization), but no longer making
the impoverished features available at the leaves (e.g.,~for
vocabulary insertion). This indicates that, in a case like
the tree of \eqref{exSobl}, if $\alpha$
and $\phi$ are the features to be retained and
$\beta$, $\gamma$, $\delta$ are those targeted by
Impoverishment, it is the term
$$
\Tree[ .{$[\phi, \alpha, \beta, \gamma,\delta]$} [ .{$[\phi, \alpha]$} $\phi$ $\alpha$ ] ] 
$$
in the right-channel of the coproduct 
$$ \Tree[ .{$[\beta, \gamma,\delta]$}
  $\beta$ [ .{$[ \gamma, \delta ]$} $\gamma$ $\delta$ ] ] \otimes
\Tree[ .{$[\phi, \alpha, \beta, \gamma,\delta]$} [ .{$[\phi, \alpha]$} $\phi$ $\alpha$ ] ] $$
that we want to use in replacement of the
original morphological tree $S$ of \eqref{exSobl} rather than 
the term
$$ 
\Tree[ .{$[\phi, \alpha]$} $\phi$ $\alpha$ ]
$$
 in the left-channel of the term of the coproduct of the form
$$ 
\Tree[ .{$[\phi, \alpha]$} $\phi$ $\alpha$ ] \otimes
\Tree[ .{$[\phi, \alpha, \beta, \gamma,\delta]$} [ .{$[\beta, \gamma,\delta]$}
  $\beta$ [ .{$[ \gamma, \delta ]$} $\gamma$ $\delta$ ] ] ] \, .
$$ 
This requires 
moving a term from the right-channel of the coproduct to the 
workspace and then to the left-channel of a second application
of the coproduct. This can be achieved, for morphological 
workspaces by first acting with the simplest Hopf algebra Markov chain $\sqcup \circ \Delta^\rho$. Applied to a workspace $F$,
it generates a sum of terms, one of which is of the form
$S_{B_2} \sqcup S_1 \sqcup \cdots \sqcup S_n \sqcup S/^\rho S_{B_2} $, with $S_{B_1}, S_{B_2}$ as in \eqref{SB1SB2ex}, for $F=S \sqcup S_1 \sqcup \cdots \sqcup S_n$. 

(A similar situation arises the case of Internal Merge in syntax,
where the extracted $T_v$ is first deposited in the workspace and then merged with $T/T_v$.)

 We can formalize the procedure described above in the following way.

 \begin{prop}\label{ImpovProp}
 The impoverishment operations $\bI_{B\subset B_v}: \cV(\cW_{\cM\cS}) \to \cV(\cW_{\cM\cS})$ and 
 $\bI_{B_v/B}: \cV(\cW_{\cM\cS}) \to \cV(\cW_{\cM\cS})$
 have two cases:
 \begin{enumerate}
 \item Obliteration of a subbundle of features (as discussed in \S \ref{OblSubSec}).
 \item Obliteration of a subbundle of features that maintains their trace (as outlined above).
 \end{enumerate}
In the first case, for $B\subset B_v$ at the root $v$ of an extended morphological tree $S$, the operation $\bI_{B\subset B_v}$
acts as
\begin{equation}\label{Imp1}
\bI_{B\subset B_v}(\gamma_{\cS\cO,\cM\cO}(T,S,S_1,\ldots, S_n))=
\gamma_{\cS\cO,\cM\cO}(T,S_{B'},S_1,\ldots, S_n),
\end{equation}
where $B_v=B\sqcup B'$ and $S_{B'}$ is the subtree of the fission of $S$ according to $B_v=B\sqcup B'$.

This impoverishment operation can be equivalently described as mapping
\begin{equation}\label{Imp2}
\bI_{B\subset B_v}: \fM^T_{S,S_1,\ldots,S_n} \mapsto \fM^T_{S_{B'},S_1,\ldots,S_n} \, .
\end{equation}
The second case is similar, but of the form
\begin{equation}\label{Imp3}
\bI_{B_v/B}(\gamma_{\cS\cO,\cM\cO}(T,S,S_1,\ldots, S_n))=
\gamma_{\cS\cO,\cM\cO}(T,\cF_v\Phi_{A, (B,B')}/S_{B\cup A},S_1,\ldots, S_n),
\end{equation}
where $\cF_v\Phi_{A, (B,B')}$ is a fission of $S$ with
$S_{B\cup A}$ and $S_{B'\cup A}$ the two fissioned subtrees,
followed by fusion $\cF_v$ at its syntactic root vertex $v$.
This is equivalent to the formulation
\begin{equation}\label{Imp4}
\bI_{B_v/B}: \fM^T_{S,S_1,\ldots,S_n} \mapsto \fM^T_{\cF_v\Phi_{A, (B,B')}/S_{B\cup A},S_1,\ldots,S_n} \, .
\end{equation}
\end{prop}

Finally, we observe that the Obliteration and
Impoverishment operations are not additional
independent operations, but are obtainable from
the fission and fusion operations and the
coproduct $\Delta^\rho$. Thus, the basic
DM operations can be reduced to just
fusion and fission.

\begin{prop}\label{OblImpfollow} 
The operations of Obliteration and
Impoverishment are obtainable from
combinations of fission, fusion, and
the coproduct $\Delta^\rho$.
\end{prop}

\proof The case of Obliteration of a full feature
bundle is already described in the proof of
Proposition~\ref{oblitprop} in terms of the
primitive part of the coproduct, so we focus on
Impoverishment. For $B_v\smallsetminus A=B\sqcup B'$ will 
use the shorthand notation for the fission operation:
$$
S \mapsto \fM_\Phi^{\rm morph}(S_{B\cup A},S_{B'\cup A}):= \gamma_{\cS\cO,\cM\cO}(\fM(\alpha_v,\alpha),S_{B\cup A},S_{B'\cup A}) \, ,
$$
and for the composition of a fission and a fusion
$$ 
S \mapsto \fM_{\cF\Phi}^{\rm morph}(S_{B\cup A},S_{B'\cup A}):= \cF_v(\gamma_{\cS\cO,\cM\cO}(\fM(\alpha_v,\alpha),S_{B\cup A},S_{B'\cup A})) \, ,
$$
where $\cF_v$ denotes the term of the fusion
operation $\cF$ that applies at the root vertex
$v$ of $\fM(\alpha_v,\alpha)$, which acquires
the label $(B_v,\alpha_v)$.

Note that this second operation, consisting of the composition of a fission and a fusion, does not in general give back $S$, 
because the two fissioned terms $S_{B\cup A},S_{B'\cup A}$
do not in general satisfy 
$S=\fM^{\rm morph}(S_{B\cup A},S_{B'\cup A})$, see 
Example~\ref{fissex2}.

The first transformation $\bI_{B\subset B_v}$ 
can be seen, in terms of
building operations acting on morphological workspaces to
assemble morphosyntactic objects, as the transformation that
starts with a morphological workspace of the form
$F= S \sqcup \sqcup_i S_i \sqcup F'$, and proceeds as
$$ F \stackrel{\Phi_S}{\longmapsto} \Phi_S(F)= \fM_\Phi^{\rm morph}(S_{B},S_{B'})\sqcup_i S_i \sqcup F' $$ $$ \stackrel{\Delta^\rho}{\longmapsto}  S_{B'}\sqcup 
\fM_\Phi^{\rm morph}(S_{B},S_{B'})/^\rho S_{B'}
\sqcup_i S_i \otimes  F' + \text{ other terms } 
$$
$$ \stackrel{\sqcup}{\mapsto} 
S_{B'}\sqcup 
\fM_\Phi^{\rm morph}(S_{B},S_{B'})/^\rho S_{B'}
\sqcup_i S_i \sqcup  F' + \text{ other terms } 
\, , $$
where we write $\Phi_S$ for a fission operation that
targets the component $S$ of the workspace $F$.
We then proceed with the new workspace
$$ \tilde F:= S_{B'}\sqcup 
\fM_\Phi^{\rm morph}(S_{B},S_{B'})/^\rho S_{B'}
\sqcup_i S_i \sqcup  F' $$
with
$$ \tilde F \stackrel{\Delta^\rho}{\longmapsto} 
( S_{B'} \sqcup_i S_i ) \otimes ( \fM_\Phi^{\rm morph}(S_{B},S_{B'})/^\rho S_{B'} \sqcup F' ) + \text{ other terms }
$$
$$ \stackrel{\gamma_{\cS\cO,\cM\cO}(T,\cdots)\otimes {\rm id}}{\longmapsto} \gamma_{\cS\cO,\cM\cO}(T,S_{B'},S_1,\ldots, S_n) \otimes ( \fM_\Phi^{\rm morph}(S_{B},S_{B'})/^\rho S_{B'} \sqcup F' )
$$
$$
\stackrel{\sqcup}\mapsto \gamma_{\cS\cO,\cM\cO}(T,S_{B'},S_1,\ldots, S_n) \sqcup \fM_\Phi^{\rm morph}(S_{B},S_{B'})/^\rho S_{B'} \sqcup \sqcup F' $$
where $$\gamma_{\cS\cO,\cM\cO}(T,S_{B'},S_1,\ldots, S_n)$$ is
the resulting morphosyntactic object formed, while 
$$\fM_\Phi^{\rm morph}(S_{B},S_{B'})/^\rho S_{B'} \sqcup F'$$ 
is the remaining discarded morphological material, that remains
available for other morphosyntactic constructions.

The second case is similar. Again starting with a 
workspace of the form
$F= S \sqcup \sqcup_i S_i \sqcup F'$ we proceed in the following way:
$$ F \stackrel{\cF_v\circ\Phi_S}{\longmapsto} \Phi_S(F)= \fM_{\cF\Phi}^{\rm morph}(S_{B\cup A},S_{B'\cup A})\sqcup_i S_i \sqcup F' 
$$
$$
\stackrel{\sqcup \circ\Delta^\rho}{\longmapsto}
S_{B\cup A} \sqcup_i S_i  \sqcup \fM_{\cF\Phi}^{\rm morph}(S_{B\cup A},S_{B'\cup A})/^\rho S_{B\cup A} \sqcup F' + \text{ other terms} 
$$
$$
\stackrel{\Delta^\rho}{\longmapsto}
(\fM_{\cF\Phi}^{\rm morph}(S_{B\cup A},S_{B'\cup A})/^\rho S_{B\cup A} \sqcup_i S_i) \otimes (S_{B'\cup A})\sqcup F') + \text{ other terms } 
$$
$$
\stackrel{\gamma_{\cS\cO,\cM\cO}(T,\cdots)\otimes {\rm id}}{\longmapsto}  \gamma_{\cS\cO,\cM\cO}(T,\fM_{\cF\Phi}^{\rm morph}(S_{B\cup A},S_{B'\cup A})/^\rho S_{B\cup A},S_1,\ldots, S_n) \otimes 
( S_{B\cup A} \sqcup F') 
$$
$$
\stackrel{\sqcup}{\mapsto} \gamma_{\cS\cO,\cM\cO}(T,\fM_{\cF\Phi}^{\rm morph}(S_{B\cup A},S_{B'\cup A})/^\rho S_{B\cup A},S_1,\ldots, S_n) \sqcup
S_{B\cup A} \sqcup F' \, ,
$$
where $\gamma_{\cS\cO,\cM\cO}(T,\fM_{\cF\Phi}^{\rm morph}(S_{B\cup A},S_{B'\cup A})/^\rho S_{B\cup A},S_1,\ldots, S_n)$
is the resulting morphosyntactic object and
$S_{B\cup A} \sqcup F'$ is the discarded morphological material
that remains available for further structure-building operations.
\endproof
 
\begin{rem}\label{insImp}{\rm
The form \eqref{Imp3}, \eqref{Imp4} of the Impoverishment
operation allows for implementing in our model the
insertion of the unmarked feature that is relevant to
various examples of Impoverishment. We will be discussing
this more in detail elsewhere.
}\end{rem}

\section{The movable boundary of morphosyntax} \label{BoundarySec}

The description of the fundamental operations of Distributed Morphology given above suggests that
we should consider them as operations acting on the set $\cL_{\cS\cO,\cM\cO}$ of \eqref{LSOMO}
of morphosyntax building operations. Indeed, the characterization of fusion and fission as given in 
Proposition~\ref{Fusediagr} and Proposition~\ref{fissdiagr} allows us to identify fusion and fission with
transformations 
$$ \cF_{\gamma_{\cS\cO,\cM\cO}(\fM(\alpha_1,\alpha_2), S_1,S_2)}: \cL_{\cS\cO,\cM\cO} \to \cL_{\cS\cO,\cM\cO} $$
\begin{equation}\label{FuLSOMO}
\cF_{\gamma_{\cS\cO,\cM\cO}(\fM(\alpha_1,\alpha_2), S_1,S_2)}:  \fM^T_{S_1,S_2,\ldots S_n} \mapsto 
\fM^{T/^c \fM(\alpha_1,\alpha_2)}_{S_{12},\ldots S_n} \, , 
\end{equation}
and 
$$ \Phi_{A,(B_1,B_2),\alpha} : \cL_{\cS\cO,\cM\cO} \to \cL_{\cS\cO,\cM\cO} $$
\begin{equation}\label{FiLSOMO}
\Phi_{A,(B_1,B_2),\alpha} : \fM^T_{S_1,\ldots, S_\ell, \ldots S_n} \mapsto \fM^{T\circ_\ell \fM(\alpha_\ell,\alpha)}_{S_1,\ldots, S_{\ell, B_1\cup A}, S_{\ell, B_2\cup A}, \ldots S_n}
\end{equation}
In a similar way, the Impoverishment or Obliteration operations of DM can
be seen (up to composition with fission operations, as in Example~\ref{exImp2} above) 
as transformations of $\cL_{\cS\cO,\cM\cO} \to \cL_{\cS\cO,\cM\cO}$ mapping
\begin{equation}\label{ImpoLSOMO}
 \fM^T_{S_1,\ldots, S_\ell, \ldots S_n}  \mapsto \fM^T_{S_1,\ldots, S_\ell' ,\ldots S_n} 
\end{equation} 
where $S_\ell' \subset S_\ell$ is one of the two accessible terms immediately below the root
of $S_\ell$.  Thus, we can view DM operations as a semigroup action on the set 
$\cL_{\cS\cO,\cM\cO}$ obtained by arbitrary compositions of the generators
\eqref{FuLSOMO}, \eqref{FiLSOMO}, \eqref{ImpoLSOMO}. 

\begin{defn}\label{DMsemi}
The {\em Distributed Morphology semigroup} 
$\cS_{DM}$ is the semigroup generated by the operations \eqref{FuLSOMO}, \eqref{FiLSOMO}, \eqref{ImpoLSOMO},
acting on the set $\cL_{\cS\cO,\cM\cO}$.  We refer to the subsemigroup generated by 
\eqref{FuLSOMO}, \eqref{FiLSOMO} as the {\em post-syntactic semigroup} $\cS_{PS} \subset \cS_{DM}$.
\end{defn}

The main effect of the action of the semigroup $\cS_{DM}$ on the set $\cL_{\cS\cO,\cM\cO}$
is to render the boundary between syntax and morphology in the construction of morphosyntactic trees
flexible, or movable (via the action of the generators \eqref{FuLSOMO}, \eqref{FiLSOMO} and the 
post-syntactic semigroup $\cS_{PS}$), along
with the possibility of dropping some morphological features as effect of the generator \eqref{ImpoLSOMO}.
We focus here on the action of the post-syntactic semigroup $\cS_{PS}$. 

\smallskip

We refer to the dynamics implemented by these transformations as post-syntactic because
it relies on formed syntactic objects and acts on the operations that interface syntax with
morphology, hence they are not part of the computational structure of syntax, rather they
are properties of the interface with morphology. 

\smallskip

In a morphosyntactic tree, the two generators \eqref{FuLSOMO}, \eqref{FiLSOMO} of $\cS_{PS}$, 
representing the fusion and fission operations, respectively move upward or downward the vertices 
where the boundary between syntax and morphology occurs. In fusion, a syntactic vertex above the
morphological insertions becomes the place where the morphological insertion takes place, while
in fission a vertex of morphological insertion at a leaf of a syntactic tree becomes an interior
syntactic vertex. In terms of the action on $\cL_{\cS\cO,\cM\cO}$  this change is achieved by
altering the assembly procedure of the morphosyntactic structure (replacing an element of
$\cL_{\cS\cO,\cM\cO}$, which is one such assembly procedure, with another one). 

\smallskip

In Externalization, a filter can restrict the range of applicability of these transformations,
limiting (in a language depended way) the amount of flexibility in the boundary between
morphology and syntax, with polysynthetic languages (like Inuktitut) at one extreme, 
where the boundary can be significantly pushed upward into syntax, effectively absorbing
syntax into word formation and morphology, and the most 
analytic languages (like Vietnamese) at the opposite extreme, where the boundary is
pushed all the way downward, and with intermediate possibilities, for example agglutinating 
languages (like Swahili), fusional languages (like Semitic languages), 
oligosynthetic (like Nahuatl). 

\subsection{Additional remarks}

\smallskip

 We briefly discuss some additional remarks about the construction presented here.
 Note that the construction of morphosyntactic objects discussed in \S \ref{MorphoSyntaxSec2} would still work if one wants to
consider more general forms of morphological trees that also involve higher valency vertices: the syntactic part of
the tree would remain binary, while the morphological parts would 
include both binary and higher valence vertices. One may worry then that the presence of non-binary trees in the morphological part would affect the possibility of moving the morpho-syntactic boundary via the fusion and fission operations, but this is not the case. The fusion operation transforms a binary syntactic
vertex into a binary morphological vertex, which would still be available. The fission operations are design to split a bundle of morphological features into {\em two} parts and construct the
two corresponding morphological trees using the set theoretic splitting of the feature bundle and the original morphological tree structure, as described in Definition~\ref{fissiondef}. This will in any case generate a new syntactic vertex that is necessarily binary, as required for syntax, even if the resulting two morphological trees produced following the algorithm of 
Definition~\ref{fissiondef} may have higher valence vertices.
Thus, the boundary between syntax and morphology would remain movable even if morphological structures are realized by trees 
with higher valence vertices (as used in the case of feature geometry). Our choice to represent all morphological 
tree as binary is motivated by optimality (significant
simplification of the algebraic structure generating them).

 \subsection*{Acknowledgment} This work is supported by NSF grant DMS-2104330, by Caltech's Center of Evolutionary Science,
and by Caltech's T\&C Chen Center for Systems Neuroscience. 
 
\newpage
\bibliographystyle{apalike}
\bibliography{references}

\begin{thebibliography}{}

\bibitem[Arregi and Nevins, 2012]{arregiMorphotacticsBasque2012}
Arregi, K. and Nevins, A. (2012).
\newblock {\em Morphotactics: {Basque} auxiliaries and the structure of
  spellout}.
\newblock Springer Dordrecht.

\bibitem[Baunaz et~al., 2018a]{baunaz_exploring_2018}
Baunaz, L., Haegeman, L., De~Clercq, K., and Lander, E. (2018a).
\newblock {\em Exploring nanosyntax}.
\newblock Oxford University Press.

\bibitem[Baunaz et~al., 2018b]{baunaz_nanosyntax_2018}
Baunaz, L., Lander, E., De~Clercq, K., and Haegeman, L. (2018b).
\newblock Nanosyntax: the basics.
\newblock {\em Oxford Studies in Comparative Syntax}, pages 3--56.
\newblock Publisher: Oxford University Press.

\bibitem[Bobaljik, 1997]{bobaljik_if_1997}
Bobaljik, J.~D. (1997).
\newblock If the head fits...: on the morphological determination of {Germanic}
  syntax.
\newblock Publisher: Walter de Gruyter, Berlin/New York Berlin, New York.

\bibitem[Caha, 2009]{caha_nanosyntax_2009}
Caha, P. (2009).
\newblock The nanosyntax of case.
\newblock Publisher: Universitetet i Tromsø.

\bibitem[Coon and Keine, 2021]{coon_feature_2021}
Coon, J. and Keine, S. (2021).
\newblock Feature gluttony.
\newblock {\em Linguistic inquiry}, 52(4):655--710.
\newblock Publisher: MIT press journals-info@ mit. edu.

\bibitem[Cowper and Hall, 2002]{cowper_syntactic_2002}
Cowper, E. and Hall, D.~C. (2002).
\newblock The syntactic manifestation of nominal feature geometry.
\newblock In {\em Proceedings of the 2002 annual conference of the {Canadian}
  {Linguistic} {Association}}, pages 55--66. Canadian Linguistic Association
  Toronto.

\bibitem[Cowper, 2005]{cowper_geometry_2005}
Cowper, E.~A. (2005).
\newblock The geometry of interpretable features: {Infl} in {English} and
  {Spanish}.
\newblock {\em Language}, 81(1):10--46.
\newblock Publisher: Linguistic Society of America.

\bibitem[Embick and Noyer, 2007]{embick_distributed_2007}
Embick, D. and Noyer, R. (2007).
\newblock Distributed morphology and the syntax—morphology interface.

\bibitem[Halle and Marantz, 1993]{halle_distributed_1993}
Halle, M. and Marantz, A. (1993).
\newblock Distributed morphology and the pieces of inflection.
\newblock In Hale, K. and Keyser, S., editors, {\em The view from building 20},
  pages 111--176. The MIT Press, Cambridge, MA.

\bibitem[Halle and Marantz, 1994]{halle_key_1994}
Halle, M. and Marantz, A. (1994).
\newblock Some key features of distributed morphology.
\newblock In Carnie, A., Harley, H., and Bures, T., editors, {\em {MITWPL} 21},
  pages 275--288.

\bibitem[Harbour, 2008]{harbour_discontinuous_2008}
Harbour, D. (2008).
\newblock Discontinuous agreement and the syntax-morphology interface.
\newblock {\em Phi theory: Phi-features across modules and interfaces}, pages
  185--220.
\newblock Publisher: Oxford University Press Oxford.

\bibitem[Harley and Ritter, 2002]{harleyPersonNumber2002}
Harley, H.~E. and Ritter, E. (2002).
\newblock Person and number in pronouns: {A} feature-geometric analysis.
\newblock {\em Language}, 78:482 -- 526.

\bibitem[Haywood and Nahmad, 1965]{haywood1965arabic}
Haywood, J. and Nahmad, H. (1965).
\newblock {\em Arabic grammar of the written language}.
\newblock Lund Humphries.

\bibitem[Hewett, 2023]{hewett_allomorphy_2023}
Hewett, M. (2023).
\newblock Allomorphy in {Semitic} discontinuous agreement: {Evidence} for a
  modular approach to postsyntax.
\newblock {\em Natural Language \& Linguistic Theory}, 41(3):1091--1145.
\newblock Publisher: Springer.

\bibitem[Kleine and M\"uller, rint]{keine_impoverishment_nodate}
Kleine, S. and M\"uller, G. (preprint).
\newblock Impoverishment.

\bibitem[Marcolli et~al., 2025a]{marcolli_mathematical_2024}
Marcolli, M., Chomsky, N., and Berwick, R.~C. (2025a).
\newblock {\em Mathematical {Structure} of {Syntactic} {Merge}: {An}
  {Algebraic} {Model} for {Generative} {Linguistics}}.
\newblock MIT Press.

\bibitem[Marcolli et~al., 2025b]{marcolli_phases_2025}
Marcolli, M., Huijbregts, R., and Larson, R.~K. (2025b).
\newblock Hypermagmas and colored operads: heads, phases, and theta roles.

\bibitem[Marcolli et~al., 2024]{marcolli_extension_2024}
Marcolli, M., Larson, R., and Huijbregts, R. (2024).
\newblock Extension {Condition} ``violations" and {Merge} optimality
  constraints.

\bibitem[Marcolli and Larson, 2025]{marcolli_theta_2025}
Marcolli, M. and Larson, R.~K. (2025).
\newblock Theta theory: operads and coloring.
\newblock {\em arXiv:2503.06091}.

\bibitem[Martinović, 2022]{martinovic_feature_2022}
Martinović, M. (2022).
\newblock Feature geometry and head splitting in the {Wolof} clausal periphery.
\newblock {\em Linguistic Inquiry}, 54(1):79--116.
\newblock Publisher: MIT press journals-info@ mit. edu.

\bibitem[McGinnis, 2007]{mcginnis_phi-feature_2007}
McGinnis, M. (2007).
\newblock Phi-feature competition in morphology and syntax.
\newblock Publisher: Oxford University Press.

\bibitem[Nevins, 2016]{nevins2016lectures}
Nevins, A. (2016).
\newblock Lectures on postsyntactic morphology.
\newblock {\em Ms. University College London}.

\bibitem[Noyer, 1997]{noyer_1997}
Noyer, R. (1997).
\newblock {\em Features, Positions and Affixes in Autonomous Morphological
  Structure}.
\newblock Garland Publishing.

\bibitem[Starke, 2009]{starke_nanosyntax_2009}
Starke, M. (2009).
\newblock Nanosyntax: {A} short primer to a new approach to language.
\newblock {\em Nordlyd}, 36(1):1--6.

\bibitem[Svenonius, 2006]{svenonius_case_2006}
Svenonius, P. (2006).
\newblock Case alternations and the {Icelandic} passive and middle.
\newblock {\em Passives and impersonals in European languages}.
\newblock Publisher: Citeseer.

\bibitem[Tosco, 2007]{tosco_feature-geometry_2007}
Tosco, M. (2007).
\newblock Feature-geometry and diachrony: the development of the subject
  clitics in {Cushitic} and {Romance}.
\newblock {\em Diachronica}, 24(1):119--153.
\newblock Publisher: John Benjamins.

\bibitem[Watson, 1993]{watson1993syntax}
Watson, J.~C. (1993).
\newblock {\em A syntax of Sanani Arabic}, volume~13.
\newblock Otto Harrassowitz Verlag.

\bibitem[Xu, 2018]{xu_double-person_2018}
Xu, Y. (2018).
\newblock A double-person feature analysis of {Algonquian} first person
  plurals.
\newblock In {\em Proceedings of {WCCFL}}, volume~35.

\end{thebibliography}

\end{document}